\documentclass[journal]{acmart}

\usepackage{cprotect}
\usepackage{subcaption}
\usepackage[linesnumbered,ruled]{algorithm2e}
\usepackage{multirow}

\usepackage{array}
\usepackage{ragged2e}
\newcolumntype{P}[1]{>{\RaggedRight\hspace{0pt}}p{#1}}
\usepackage{color, colortbl}
\definecolor{Gray}{gray}{0.9}
\usepackage{savesym}
\savesymbol{comment}
 \usepackage[final]{changes}
\restoresymbol{CHNG}{comment}
\setaddedmarkup{\color{blue}{#1}}
\setdeletedmarkup{\color{red}{\sout{#1}}}

\AtBeginDocument{%
  \providecommand\BibTeX{{%
    \normalfont B\kern-0.5em{\scshape i\kern-0.25em b}\kern-0.8em\TeX}}}

\acmJournal{THRI}

\begin{document}

\title{SocNavBench: A Grounded Simulation Testing Framework for Evaluating Social Navigation}

\author{Abhijat Biswas}
\email{abhijat@cmu.edu}
\author{Allan Wang}
\author{Gustavo Silvera}
\author{Aaron Steinfeld}
\author{Henny Admoni}
\affiliation{%
  \institution{Carnegie Mellon University}
  \streetaddress{5000, Forbes Avenue}
  \city{Pittsburgh}
  \state{Pennsylvania}
  \postcode{15213}
}

\renewcommand{\shortauthors}{Biswas, et al.}

\begin{abstract}
The human-robot interaction (HRI) community has developed many methods for robots to navigate safely and socially alongside humans. However, experimental procedures to evaluate these works are usually constructed on a per-method basis. Such disparate evaluations make it difficult to compare the performance of such methods across the literature.
To bridge this gap, we introduce \emph{SocNavBench}, a simulation framework for evaluating social navigation algorithms. \emph{SocNavBench} comprises a simulator with photo-realistic capabilities and curated social navigation scenarios grounded in real-world pedestrian data. We also provide an implementation of a suite of metrics to quantify the performance of navigation algorithms on these scenarios. Altogether, \emph{SocNavBench} provides a test framework for evaluating disparate social navigation methods in a consistent and interpretable manner. 
To illustrate its use, we demonstrate testing three existing social navigation methods and a baseline method on \emph{SocNavBench}, showing how the suite of metrics helps infer their performance trade-offs.
Our code is open-source, allowing the addition of new scenarios and metrics by the community to help evolve \emph{SocNavBench} to reflect advancements in our understanding of social navigation.
\end{abstract}



\maketitle

\section{Introduction}

\begin{figure}[htbp]
    \centering
    \includegraphics[width=\textwidth]{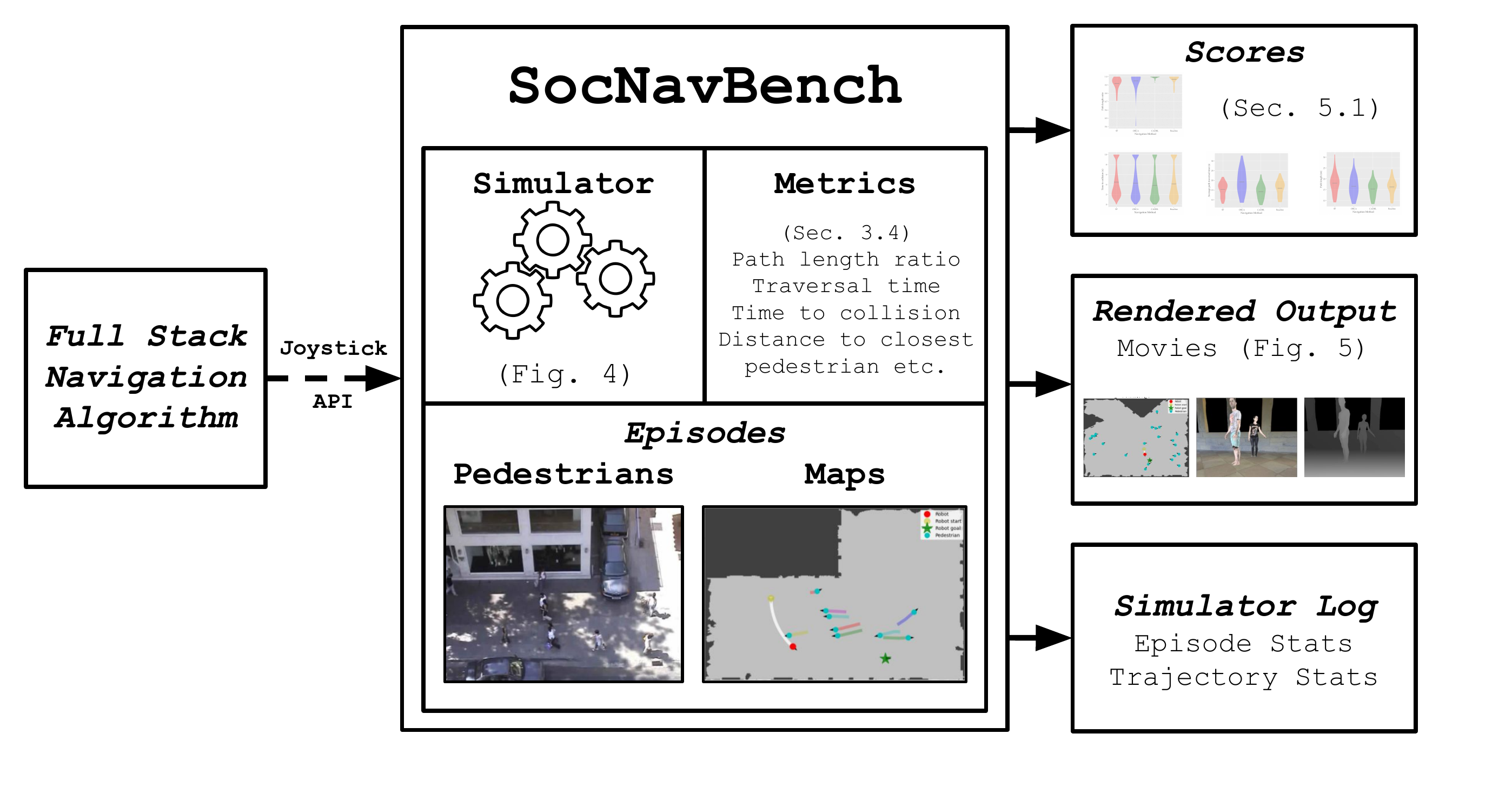}
    \caption{SocNavBench is a social navigation test bench comprising of a photo-realistic renderer, a curated set of navigation scenarios based on real-world pedestrian data, and a suite of metrics to characterize the performance of robot navigation algorithms. SocNavBench can take as input a full or partial stack navigation algorithm. It supports simulated photorealistic RGB-D sensing and model based control.} 
    \label{fig:sim_overview}
\end{figure}


Robots must navigate in a safe, predictable, and socially acceptable manner in order to succeed in spaces designed for and occupied by humans -- often referred to as social navigation. Many methods for social navigation have emerged in the human-robot interaction (HRI) literature \cite{helbing1995social, trautman2010unfreezing, kirby2010social, luber2012socially, vandenberg2008, vasquez2014inverse_sim, chen2017socially, mavrogiannis2017socially, mavrogiannis2018social}
focusing on large dense crowds \cite{helbing1995social, trautman2010unfreezing}, local interactions with small crowds \cite{mavrogiannis2018social, vandenberg2008}, and specific methods for service scenarios \cite{burgard1993Rhino, thrun99minerva, mutlu2008robots}. However, it remains difficult to evaluate these methods against one another in a consistent manner. Each algorithm tends to be evaluated using scenarios and metrics chosen by the researchers for their specific implementations. There is currently no single method that enables the characterization of the strengths and weaknesses of different social navigation approaches in a consistent and interpretable manner. 
In this work, we introduce a simulation-based evaluation framework to enable direct comparisons between different social navigation approaches. Our algorithm-agnostic framework uses ecologically valid test scenarios based on real-world pedestrian data and introduces a series of metrics that quantify different aspects of navigation.

Social navigation is a difficult task to benchmark for several reasons. 
First, what constitutes successful social navigation is subjective and heavily dependent on contextual factors including location, available space, the mobility of surrounding humans, the robot's function, and local or cultural norms.
For example, it is much more acceptable for a robot to cut across pedestrian paths if ferrying emergency medicine to a mall customer experiencing an anaphylactic reaction than if delivering food to a customer in the food court. However, not all scenarios are this clear.
The same robot performing the same task in a hospital should now avoid obstructing nurses, doctors, and patients.
In our navigation benchmark, we provide a number of automatically computed metrics that quantify system characteristics, such as robot energy expenditure, path smoothness, \deleted{legibility,} speed, and safety, among others. Using these metrics, evaluations of social navigation algorithms can prioritize characteristics appropriate for the robot's context.



Second, it is difficult to create consistent and repeatable realistic navigation scenarios with which to fairly compare different algorithms, while also being faithful to real-world pedestrian behavior. If we attempted to replicate scenarios across different robots and algorithms in the real world, we would have to artificially constrain human participants to follow certain paths or move to fixed targets. However, such artificially contrived scenarios can cause people to behave in unnatural and constrained ways, which would render such an evaluation ineffective at representing real-world, natural social navigation. Instead, our navigation benchmark evaluates social navigation in simulation, allowing for perfectly repeatable experiments. To keep our simulator representative of real-world pedestrians, we crafted $33$ representative navigation scenarios with real-world pedestrian trajectories. Additionally, the simulator provides 3D rendering and a depth map for algorithms that use those in order to provide a more realistic simulation.

Third, like most HRI problems, social navigation is subject to immense variation amongst humans. 
While many social navigation algorithms are currently evaluated in simulation
\replaced{\cite{helbing1995social,vandenberg2008,chen2017socially,mavrogiannis2016decentralized, treuille2006,karamouzas2012,chen2019,duToit2012,vasquez2014inverse_sim,henry2010}}{\cite{mavrogiannis2016decentralized}}
, such simulators are usually constructed ad-hoc by each individual research group, with pedestrian trajectories generated via models of pedestrian behavior. 
Evaluating with synthetically constructed pedestrian trajectories means that the evaluation may not represent real-world performance around human pedestrians. In contrast, our simulator uses real-world pedestrian data replayed in realistic navigation scenarios. The data are sourced from several open-source data sets that capture a variety of crowd characteristics and navigation environments. 
While this does not account for mutual motion where pedestrians react to the robot motions, this does support the default social navigation goal of not affecting the humans' intended motion. 

In summary, our work addresses existing gaps in the evaluation of social navigation algorithms with \emph{SocNavBench}, a pre-recorded pedestrian simulation framework. We include a set of curated episodes containing a variety of social navigation scenarios in different environments and featuring different crowd characteristics. Further, we propose a suite of metrics that evaluate social navigation algorithms along complementary axes of performance to illustrate the trade-offs that social navigation algorithms must make. These trade-offs include balancing robot speeds and directness of trajectories with causing minimal pedestrian disruption. 

SocNavBench is useful for researchers interested in comparing social navigation algorithms and robot designers interested in selecting the best navigation algorithm for their application. We provide an API that supports testing any social navigation algorithm on a simulated robot in one of 33 scenarios (Figure \ref{fig:sim_overview}). SocNavBench automatically outputs metric scores, a detailed simulator log, and videos that show the robot in action. Robotics researchers can use SocNavBench to compare their own algorithms to prior work with a repeatable scenario using consistent metrics, enabling apples-to-apples comparisons.
Given several social navigation algorithms and a particular scenario in which to deploy a robot, a system designer can also use SocNavBench as a diagnostic tool to choose which algorithm  (or specific parameters for a given algorithm) is appropriate for their particular context. 
To illustrate how SocNavBench can be employed, we implement three popular social navigation algorithms \cite{helbing1995social, vandenberg2008, chen2017socially} and a naive, pedestrian-unaware baseline method within the simulator, and we report results comparing them.


In this paper, our contributions are:
\begin{enumerate}
    \item a photo-realistic simulator with a navigation algorithm agnostic API,
    \item a curated set of episodic social navigation scenarios comprising various environment layouts and pedestrian densities based on real-world pedestrian data for use with the simulator,
    \item a suite of evaluation metrics to measure the performance of social navigation algorithms across these episodes,
    \item a test bench comprising the above three components for evaluating social navigation algorithms, and
    \item a comparison of three existing social navigation strategies and a baseline using the aforementioned episodes and metrics.
\end{enumerate}

\section{Related Work}
\subsection{Social navigation methods}
The evolution of social navigation algorithms developed by the HRI research community is reflected by the increasing capabilities to model high\added{-level} context and human behavior variation. Early works on robot navigation focused on decoupled models. These models treat pedestrians as independent moving entities or simply dynamic obstacles. Some examples include the dynamic window approach \cite{fox1997DW}, randomized dynamic planning approaches using RRT \cite{lavalle2001} and velocity obstacles approaches \cite{fiorini1998}. Later, researchers realized that the biggest challenge in addressing pedestrians lies in the uncertainty of pedestrians' future trajectories. Works such as \cite{duToit2012, joseph2010, paris2007} attempted to predict future moving patterns of the dynamic entities in the environment.

Researchers later determined that interaction modeling is the key to better human behavior modeling. Models without interaction modeling often run into \textit{the freezing robot} problem \cite{trautman2010unfreezing}. One of the earliest works to address interaction in navigation was the social forces model \cite{helbing1995social}. This model uses forces to steer agents away from obstacles and other agents and to steer agents towards goals and other attractive entities. More recent attempts to model interactions have been diverse. Vandenberg \textit{et al}. proposed reciprocal velocity obstacles to account for the reciprocity of planning under velocity obstacles for a pair of pedestrians \cite{vandenberg2008}. More recently, topology concepts have been employed in modeling how pedestrians unanimously reach a common meta-level passing strategy in a game theoretic setting \cite{mavrogiannis2016decentralized}. Trautman \textit{et al}. attempted to model interaction via a joint density term inside the pedestrians' Gaussian process mixtures models \cite{trautman2017sparse}. Other works explicitly modeled certain aspects of interaction and incorporated them in navigation, such as grouping considerations \cite{moussaid2010group,karamouzas2012,yi2015}, proxemics \cite{truong2016,bera2017socsense} and personality traits\cite{bera2017socsense}.

Growing in popularity, researchers have also attempted to use learning-based models \cite{luber2012socially, trautman2017sparse} to capture context and interaction. Deep reinforcement learning has been used to develop a collision avoidance policy, augmented with social awareness rules to inject interaction components \cite{chen2017socially, chen2019}. The model's reward definition can be adjusted to retrain policies for new context. Inverse reinforcement learning techniques \added{have also been used} in an attempt to implicitly capture interaction into a cost function by observing how real pedestrians navigate \cite{okal2016,kretzschmar2016socially,kim2016}. In a more well-defined problem space, \replaced{the pedestrian trajectory prediction problem shares a similar necessity of modeling pedestrian interactions. In this problem domain, the learning-based models contain interaction modeling modules to enhance pedestrian future states predictions, which can then be potentially adopted as future obstacle space in a navigation setting.}{pedestrian trajectory predictions have also produced many deep learning-based models that contain techniques to model context and interactions.} For example, Social-LSTM \cite{alahi2016slstm} and Social-GAN \cite{gupta2018sgan} used social pooling layers to summarize pedestrian spatial distribution for both context and interaction modeling. SoPhie \cite{sadeghian2019} additionally used top-down view image patches around pedestrians to better understand the pedestrians' surrounding context. Another interaction modeling technique used graph-based relationships \added{\cite{Mohamed2020}}\cite{vemula2018} to implicitly learn attention weights for all pairs of pedestrians, with higher weights corresponding to more interactions.



\subsection{Evaluating social navigation}
High variation in context and human behavior is not only a challenge for social navigation algorithms, but also for evaluating these algorithms. Furthermore, what set of metrics define a successful social navigation is subjective and difficult to define. Given infinite resources, the ideal test strategy could be running thousands of trials of a robot navigating through crowds in many different real-world locations. Likewise, the ideal metric could be to ask the hundreds of thousands of pedestrians who have interacted with the robot, as well as the people who tasked the robot, to rate their experiences. However, the cost of running such a study is impractical, so researchers have used various testing strategies and metrics to approximate \emph{the ideal test}.

One group of strategies to approximate \emph{the ideal test} is to evaluate social navigation algorithms in a small-scale, real-world setting. One such approach is to run a qualitative demonstration in the wild without objective or subjective metrics \cite{kretzschmar2016socially, chen2017socially}. These demonstrations are not easily reproducible and cannot be effectively used to compare different algorithms fairly. Another \deleted{, more scientific} approach is to run user studies in a controlled setting. Typically, this is done by asking a few participants to start at specific locations and reach specific goals while the robot navigates nearby \cite{kretzschmar2016socially, okal2016, mavrogiannis2019effects, truong2017toward, kirby2010social, kruse2012}. Similarly, some teams ask human participants to teleoperate a robot and compare the algorithm's performance with the humans' \cite{kim2016, trautman2015}. While these controlled, real-world test strategies offer the benefit of realism, human participant studies at scale are very expensive. This also creates limits on crowd size and context variety. It may seem possible to overcome this by conducting user studies in the wild \cite{kim2016, trautman2015, foka2010, shiomi2014}, but it is nearly impossible to repeat the same scenario for algorithm comparisons and controlled iterative development.

Another method of approximating \textit{the ideal test} is to use real-world datasets. This is accomplished by replacing one of the pedestrians with a robot and comparing the performance differences by treating the replaced pedestrian's trajectory as the ground truth trajectory \cite{trautman2010unfreezing, luber2012socially}. A similar method is to put a virtual robot in the scene and ask it to navigate to designated waypoints \cite{cao2019, bera2017socsense, bera2018socially}. The benefit of the virtual-robot-in-dataset testing strategy is that it is a good approximation to real-world scenarios as it uses real-world pedestrian trajectories and environments. Because datasets typically contain hundreds of trajectories, this testing method offers large quantities of test cases. By combining scenes from different datasets, we can further gain a decent variety of context. For these reasons, we developed our approach around real-world pedestrian trajectories in our simulator. An unmet need in prior implementations of this method is that it is impossible for a simulated robot to obtain realistic sensor inputs. Thus, the social navigation algorithms being tested need to rely on the assumption of perfect perception.

The final popular approach to approximate \textit{the ideal test} is to use simulators. Simulators offer the benefit of generating \replaced{large}{infinite} numbers of test cases, but a critical concern is what models should be used to simulate pedestrian behavior. One approach some researchers used is called "self-play" \cite{helbing1995social, vandenberg2008, chen2017socially, mavrogiannis2016decentralized, treuille2006, karamouzas2012}. Each agent in the simulated environment uses the same social navigation algorithm to ``play" against each other. Another common approach is called ``against-all" \cite{chen2019, duToit2012, vasquez2014inverse_sim, henry2010}. In this case, the social navigation algorithm is implemented on one agent. Then, that agent is tested against all other agents governed by a different model. 
The fundamental problem with these two pedestrian modeling approaches is that instead of testing against actual human behavior, the social navigation models are treated as ground truth.

Instead of simulating pedestrian behavior, we directly source real-world pedestrian trajectories from natural behavior datasets. To overcome the inability to obtain sensor inputs for these datasets, we simulate environments by applying real-world textures on pedestrians and environments, which we then use to generate photorealistic sensor inputs. Additionally, since ours is a simulated setting, we can still generate large quantities of test cases. By combining the benefits of the dataset and simulator based testing strategies, every element in our approach is grounded in reality. Therefore, we believe our testing strategy is a better approximation of \textit{the ideal test}.

In terms of the human rating metric used in \textit{the ideal test}, it is possible to ask human observers to rate how simulated robots behave in our simulators, as is the case in \cite{shiomi2014,manso2019socnav1}. However, our aim is to make our simulator a diagnostic tool that automatically outputs scores, so human ratings are not logistically reasonable. Therefore, we decided to adopt approximations to the human rating metric, as seen in most dataset- and simulator-based testing methods. Common approximations to the pedestrian experience ratings (i.e., social ratings) include collisions or success rate \cite{chen2017socially, chen2019, okal2016, bera2017socsense}, closeness to pedestrians \cite{trautman2010unfreezing, kim2016, trautman2017sparse, kretzschmar2016socially, bera2017socsense}, path predictability \cite{mavrogiannis2019effects, okal2016}, among others. Common approximations to evaluation by robot task-givers (i.e, task efficiency) include time to the reach goal \cite{okal2016, chen2017socially, chen2019, kim2016}, speed \cite{trautman2017sparse, kretzschmar2016socially}, path length \cite{trautman2010unfreezing, luber2012socially,okal2016} and other navigation performance-based metrics. Our large selection of metrics offers users of \emph{SocNavBench} the ability to prioritize metrics they deem most suitable for their specific needs. We demonstrate the use and interpretation of these metrics by testing on three different existing algorithms and providing a comparison of these metrics.

\section{SocNavBench design}
\subsection{Overview}
\begin{figure}[t]
    \centering
    \includegraphics[width=\textwidth]{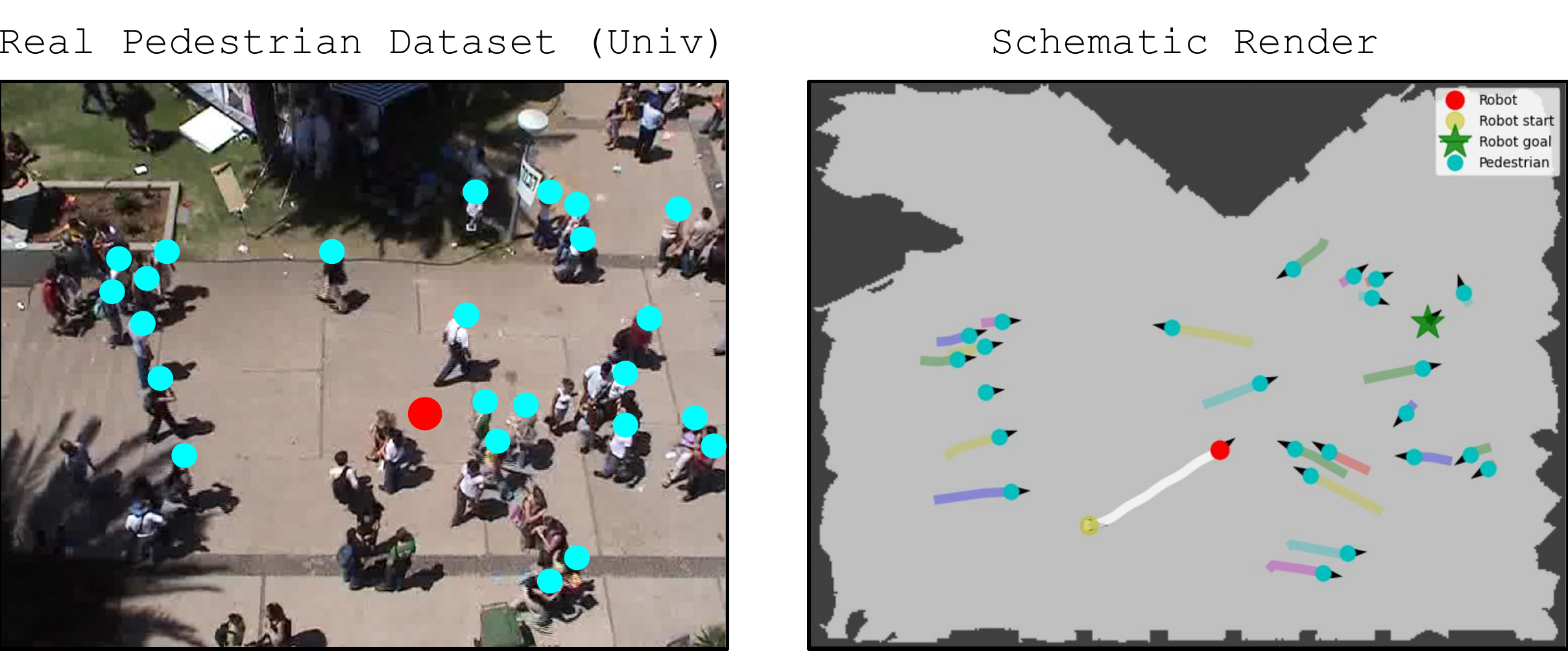}
    \caption{An example of the transfer of real data to the simulator. Pedestrian trajectories are replayed at the same speed and environment structures are faithfully reflected for navigation.}\label{fig:real2sim}
\end{figure}

Our proposed tool, \textit{SocNavBench}, is a simulator-based benchmark with prerecorded real-world pedestrian data replayed \added{(Fig. \ref{fig:real2sim})}. The simulator can run in two modes, \textit{Schematic} and \textit{Full-render}. The \textit{Schematic} mode focuses on trajectory-based navigation with the problem of perception abstracted away. In the \textit{Full-render} mode, we provide an RGB-D image with customizable camera position and intrinsic parameters to simulate real-world, on-board sensing. 

Algorithms evaluated by the tool see several episodes comprising different prerecorded human trajectories and environments. For each such episode, the algorithm under evaluation receives \replaced{one of several}{an} environment map\added{s} \added{(Fig.  \ref{fig:maps})} and a start and goal configuration within that environment that it must traverse in an efficient and socially acceptable manner. We measure robot navigation efficiency by measuring the robot's smoothness and directness of path as well as its energy expenditure.
We measure social acceptability from potential disruptions to recorded human paths, with the most socially acceptable paths not only avoiding collisions with pedestrians, but also maintaining a comfortable distance from pedestrians whenever possible. 

An assumption we use is that the pre-recorded pedestrian trajectories that were recorded in the absence of a robot are the preferred trajectories of these pedestrians. At minimum, a social robot present in the original scenario should not cause pedestrians to deviate from this path. Therefore, we should not see collisions with the pre-recorded pedestrians. 
Likewise, any robot action that leads to motions very close to replayed human trajectories would probably have induced pedestrian deviations from their preferred path. This is also sub-optimally social. \added{For a discussion of the validity and limitations of this assumption see Sec.\  \ref{sec:replay_assumptions}}

Our metrics suite is designed with these considerations in mind. In particular, we measure pedestrian disruption in two ways: a time-to-collision measure and a closest-pedestrian distance measure. The distribution of these metrics helps quantify how close the robot comes to cutting off pedestrians during its navigation. Additionally, we provide metrics that quantify the directness (vs.\ circuitousness), smoothness (vs.\ jerkiness), and energy efficiency of the robot's navigation. Together, these represent a characterization of both the appropriateness and efficiency of the robot's navigation. 



\subsection{Pre-recorded \added{pedestrian} data and episode curation} 
\label{sec:curated_eps}

\begin{figure}[t]
    \centering
    \includegraphics[width=\textwidth]{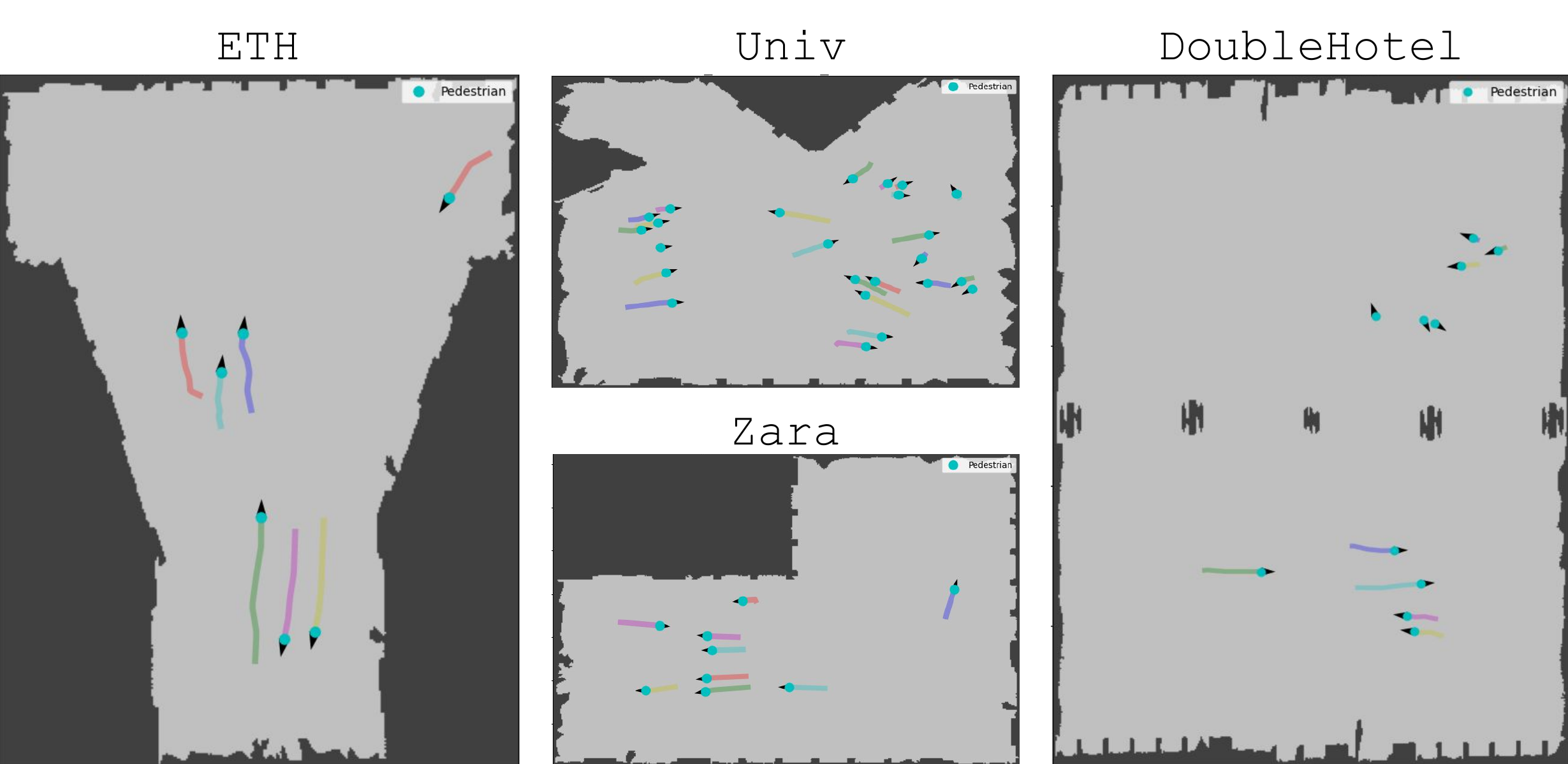}
    \caption{Example maps available in \emph{SocNavBench}. Maps have varied environment structures as well as varied crowd sizes \& densities.}\label{fig:maps}
\end{figure}

The pre-recorded pedestrian data used in SocNavBench comes from the UCY \cite{lerner2007crowds} and ETH \cite{pellegrini2009you} pedestrian trajectory datasets, which are widely accepted and used by the community \cite{cao2019, bera2017socsense, trautman2010unfreezing, alahi2016slstm, gupta2018sgan} \added{and were also included in recent pedestrian trajectory prediction benchmarks \cite{Becker2018, Kothari2021}}. These data include varied crowd densities and walking speeds, as well as interesting pedestrian behaviors, such as grouping, following, passing, pacing, and waiting. \added{Pedestrian trajectory data is replayed at the simulator tick rate (default $25$ fps) with a 1:1 ratio to recorded time, i.e. replayed pedestrian trajectories have the same velocities as the recorded ones. If the simulated robot collides with a pedestrian, SocNavBench registers that collision and the pedestrian's replayed motion is unchanged. Future support for reactive pedestrians is planned, which has its own set of assumptions and tradeoffs (see Sec.\ \ref{sec:replay_assumptions} for a discussion)}

We \replaced{manually curated a set of}{curated our} episodes \added{that include challenging scenarios from the aforementioned pre-recorded datasets}. \added{These episodes consist of} a diversity of crowd sizes, speeds, and densities, as well as directions of motion with respect to the robot's traversal \added{task, while excluding trivial scenarios with very low crowd densities}\footnote{Here, crowd density refers to the number of pre-recorded pedestrians per unit area}. We also include multiple environments (Fig. \ref{fig:maps}) with different types of configurations, hence providing environmental obstacle diversity as well. In total, we have $33$ episodes, each lasting for at most $60 s$. Our curated episodes contain an\deleted{d} average of $44$ pedestrians (standard deviation $= 13$), with the minimum being $24$ and the maximum being $72$.
\added{Additionally, to allow users access to large numbers of episodes, we have added support for random episode sampling which samples a queried number of random robot start and goal pairs in a randomly selected map along with one of the corresponding handpicked sections of recorded pedestrians from the aforementioned curated episodes. These random start and goal pairs are picked in a way that ensures they are reachable in $25s$ when using the maximum permitted robot velocity. While forgoing the benefits of hand-picked starts and goals (such as scenarios that require the robot to cross a high density crowd flow), this will allow users to use the simulator for learning purposes where large sample sizes are especially important.} 

We adapted meshes from the Stanford Large Scale 3D Indoor Spaces Dataset (SD3DIS) \cite{armeni20163d} to provide the physical environments in which these episodes to take place, modifying them to reflect the spaces in which the original data was collected. We chose a variety of segments from the pedestrian datasets to include a distribution of crowd sizes, crowd densities, pedestrian speeds, and pedestrian trajectories. Robot start and goal positions within these scenarios were sampled such that they were semantically meaningful whenever possible. For example, we chose the start position of the robot to be the exit from a building and the goal to be the end of a sidewalk in one scenario.

For perspective rendering, we based our system upon The HumANav Dataset\cite{tolani2020visual}, which uses Google’s Swiftshader renderer for photorealistic RGB and depth visuals. Importantly, the rendering engine for this project is separate from the specific meshes used, so any custom meshes could be used. The building mesh scans in the environment originated from the SD3DIS \cite{armeni20163d}, but they were modified to replicate the environments in which the pedestrian data was recorded (Fig. \ref{fig:real2sim}). Human meshes were drawn from the SURREAL Dataset \cite{varol17_surreal}, which includes photorealistic meshes for various human body configurations, genders, and lighting conditions. \added{In particular, 6000 human models form SURREAL are used that are parameterized by body configuration, pose, and velocity during walking. For replaying human poses accurately, we can calculate the body orientation and walking velocity of the recorded pedestrians from their recorded trajectory. These are then plugged into the aforementioned parameterized human models to be rendered such that accurately represent the poses that would have occurred naturally when walking at that speed and orientation.}

\subsection{SocNavBench simulator mechanics}
\begin{figure}[t]
    \centering
    \includegraphics[width=\textwidth]{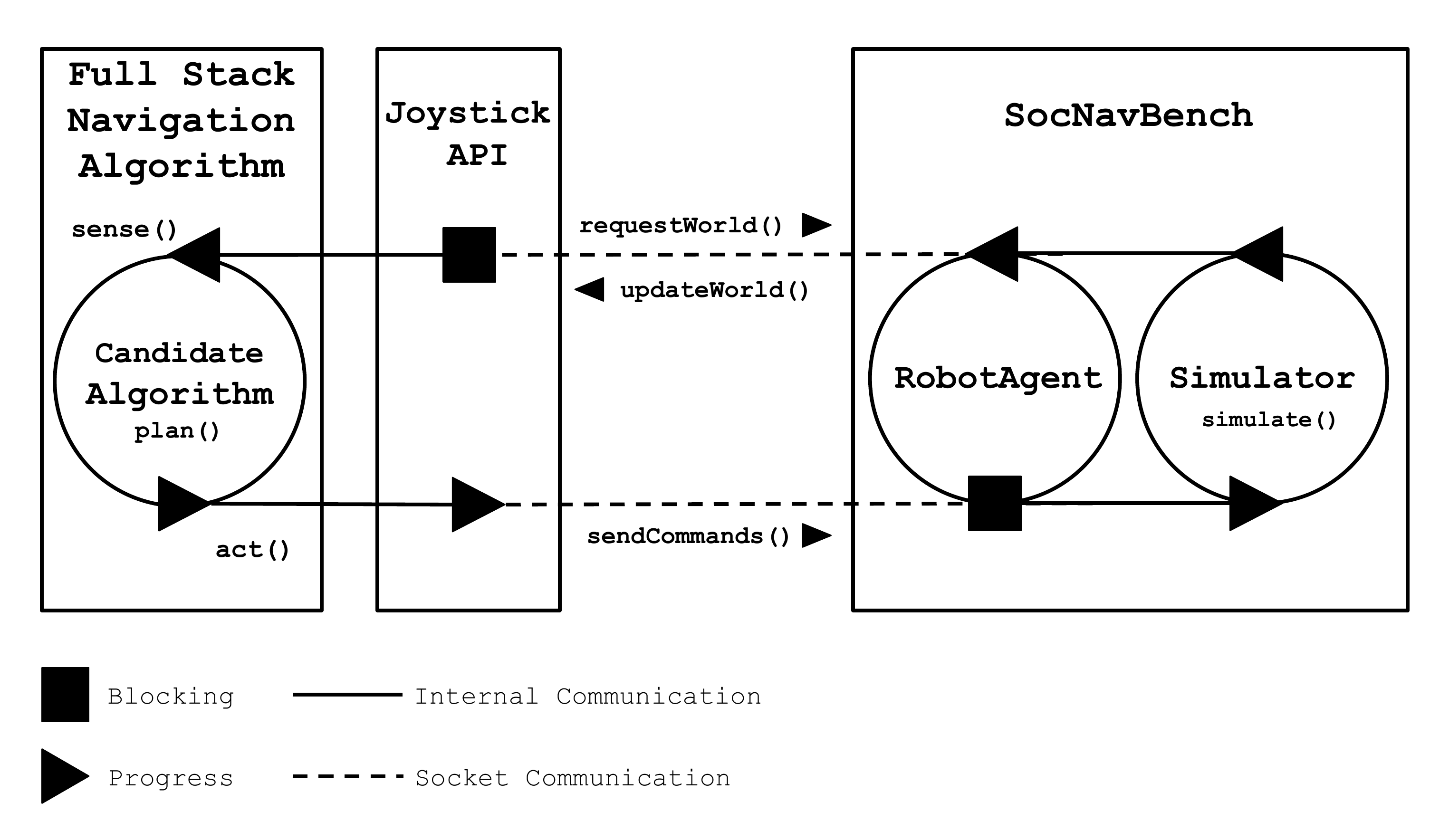}
    \cprotect
    \caption{An illustration of simulator flow control in \emph{synchronous} mode. At each iteration, the candidate algorithms uses the \verb+Joystick API+ to send a \verb+sense()+ signal, plans with the newly received world data, and then sends back an \verb+act()+ signal.}
    \label{fig:sim_SPA_mech}
\end{figure}

\emph{SocNavBench} provides a socket-based interface for candidate algorithms (clients) to communicate with the simulator (server) and the simulated RobotAgent (\added{Fig.} \ref{fig:sim_SPA_mech}). At each evaluation start, the client is sent metadata about the episodes on which it is going to be evaluated. Then, episodes are served to the client sequentially, including environment information, a time budget for each episode, and start \& goal locations. Each episode ends either when the robot reaches the goal (\textit{Completion}), collides with an environmental obstacle (\textit{Environment Collision}), or runs out of time for that episode (\textit{Timeout}). Each \textit{Completion} may be a successful episode (\textit{Success}) if there were no collisions. If the robot collided with a pedestrian during its completion of the episode, however, we count the episode as a \textit{Pedestrian Collision} and the episode cannot count as an overall success.

The flow control in \emph{SocNavBench} \replaced{can be understood}{is} based on a \added{classical} sense-plan-act cycle design \deleted{, as is typical for most robots}. When a candidate algorithm performs a sense action, it requests an updated WorldState from the simulator. This state is a position for all pedestrians in the scene in \textit{Schematic} mode and an RGB-D image in the \textit{Full-render} mode. It can then plan on the received state and send either velocity or position commands to update the simulated RobotAgent. Importantly, this abstraction allows many different types of algorithms to be compatible with \emph{SocNavBench}, including end-to-end learning based methods. This also allows individual parts of a modular navigation algorithm to be tested in an ablation study. For example, a planning algorithm could be tested with an RGB-D-based perception method in \textit{Full-render} mode and also with perfect perception using the \textit{Schematic} mode, hence isolating the navigation performance differences attributable to the perception method.

\subsubsection{Synchronicity:} 
Our simulator can run in two modes with respect to robot client and simulator server synchronicity: \textit{Asynchronous} and \textit{Synchronous}. In asynchronous mode, the simulator constantly updates at a fixed rate (simulator time is moving forward) while listening to the robot client for inputs in the background. In this mode, more complex algorithms that run slower will loop through their sense-plan-act cycles in a manner close to the real world, where taking too long to plan may result in the sensed data being stale by the time the action is performed. 

In contrast, in the synchronous mode, the simulator listens for a robot command at each simulator step. This is the default mode, which does not penalize algorithms for ``thinking time''. The synchronous option is preferred since thinking time may be dictated in part by the compute hardware used, leading to irreproducible results. However, a user may want to run the simulator in asynchronous mode for a more realistic evaluation. This is especially relevant if the user (1) can control for the computational hardware across different algorithms and (2) is testing on the hardware to be used during deployment.

\subsubsection{Rendering modes}
\begin{figure}[t]
    \centering
    \includegraphics[width=\textwidth]{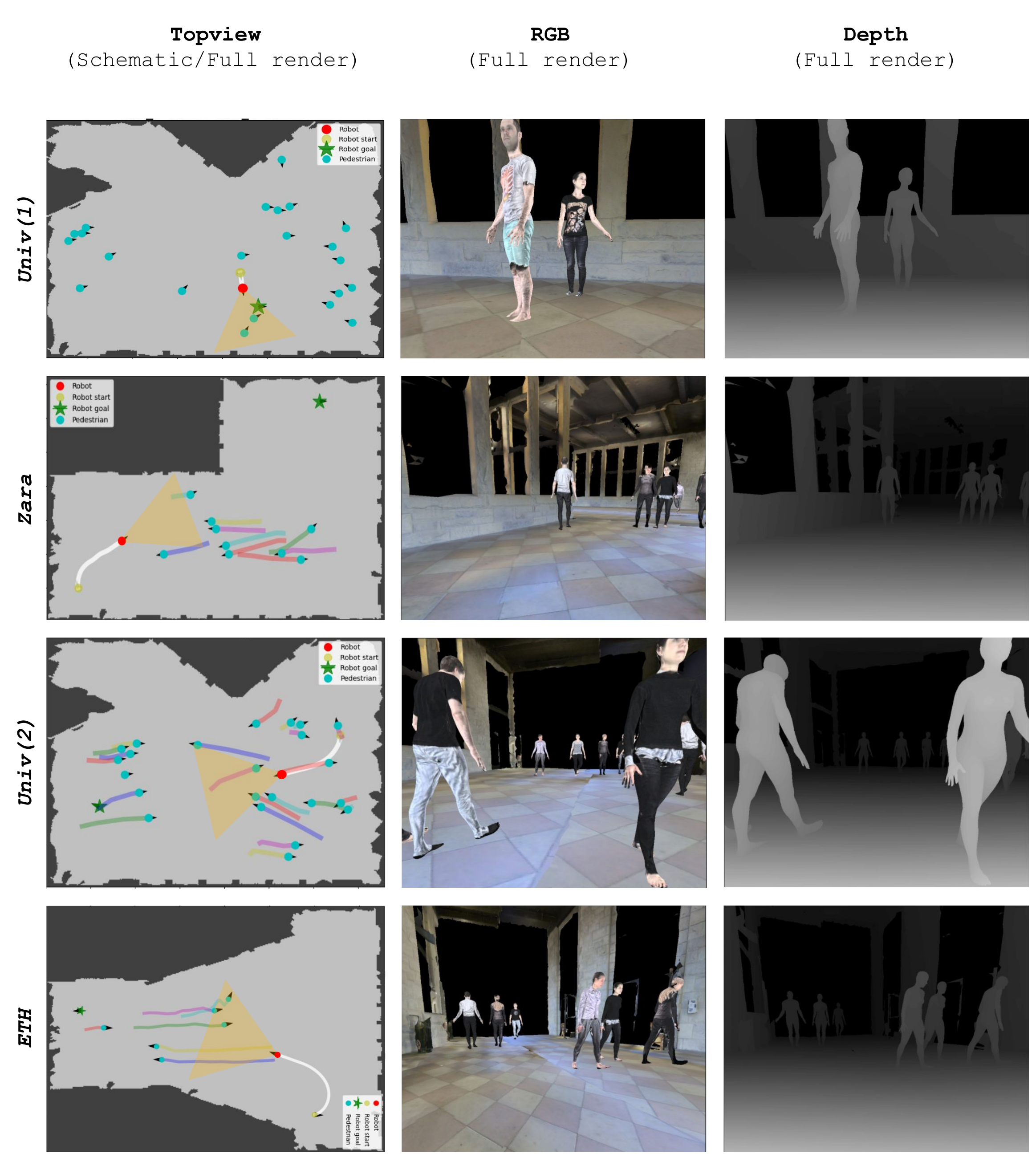}
    \cprotect
    \caption{The different types of sensing data available across both rendering modes, \textit{Schematic} and \textit{Full-render}. Camera direction is represented by the colored cone.}
    \vspace{0.3cm}
    \label{fig:render_modes}
\end{figure}

\emph{SocNavBench} supports two rendering modes: \textit{Schematic} and \textit{Full-render} (Fig. \ref{fig:render_modes}). For visualization purposes, our approach plots a schematic top-view of the entire scene, but it also includes the option of a full render from the perspective of a camera \replaced{specified by an arbitrary 6-dof pose}{placed at an arbitrarily selected height}. The default \replaced{pose}{location} corresponds to a camera attached atop a Pioneer robot \deleted{spec} \cite{pioneer-p3dx}. The \textit{Full-render} mode additionally contains building scans for the environment, and human meshes for the pedestrians. The default modes of operation are \textit{Synchronous} and \textit{Schematic}. Henceforth, when not specified, these are the modes used.


\subsubsection{Simulated robot specifications:}\label{sec:sim_bot_spec}

The simulated robot is able to run in velocity control mode and in direct position control mode. 
In velocity control mode, the robot is subject to kinematic constraints according to a three dimensional unicycle model with dynamics:
\begin{equation}\label{eq:dubins}
    \dot{x} = v \cos \phi, \quad
    \dot{y} = v \sin \phi, \quad
    \dot{\phi} = \omega
\end{equation}
where the position of the robot is ${(x, y)}$ and the heading is $\phi$.

However, most social navigation algorithms \cite{helbing1995social,chen2017socially, mavrogiannis2018social},  including ones that provide guarantees of collision-free navigation \cite{vandenberg2008}, do not include a dynamics model in the derivation of their guarantees. To ensure compatibility, we also provide a \replaced{velocity-limited holonomic}{position} control mode \added{which is akin to position control with upper bounded instantaneous velocity.}

We based the default simulated robot design on the Pioneer 3-DX robot \cite{pioneer-p3dx} using the base system dimensions and maximum velocity according to off-the-shelf specifications. In the full 3D rendering mode, the default camera height is typical for an on-board RGB-D camera on this base.
However, these options are customizable to represent any robot base and sensor position on that base \added{since any arbitrary 6-dof camera pose camera pose can be specified}.

\subsection{Evaluation criteria and metrics}
\label{sec:metrics}
\begin{table}[t]
    \centering
    \begin{tabular}{P{0.2\linewidth}p{0.08\linewidth}P{0.66\linewidth}}
    \toprule
    \textbf{Metric} & \textbf{Units} & \textbf{Definition}\\
    \midrule
    
    \addlinespace\rowcolor{Gray}\multicolumn{3}{c}{Meta statistics} \\\addlinespace
    Overall success rate & -- &
    the fraction of episodes successfully completed (robot reached goal position and did not collide with any pedestrians)
    \\
    Failure case\added{s}\linebreak \deleted{distribution} & -- &
    the distribution of failure cases over \textit{Timeout}, \textit{Pedestrian Collision}, and, \textit{Environment Collision}
    \\
    Total pedestrian\linebreak collisions & -- &
    the total number of pedestrian collisions across all episodes considered.
    \\
    Average planning wait time & $seconds$ &
    the average time spent by the simulator waiting for the planning algorithm per simulation step
    \\
    
    \addlinespace\rowcolor{Gray}
    \multicolumn{3}{c}{Path quality} \\\addlinespace
    Path length & $metres$  & the total distance traversed by the robot in the episode \\
    Path length ratio & -- &  the ratio of straight line distance between the end and goal to the robot's path length for any episode \\
    Goal traversal ratio & -- &
    the ratio of the robot's distance to goal at the episode end and distance to goal at the episode start \\
    Path irregularity & $radians$ &
    averaged over each point of the robot trajectory, this is the angle between the robot heading and the vector pointing to goal\\
    Path traversal time & $seconds$ & the total simulator time taken to traverse the robot's path
    \\
    
    \addlinespace\rowcolor{Gray}
    \multicolumn{3}{c}{Motion quality} \\\addlinespace
    Average speed & ${m/s}$ &
    average speed of the robot over its entire trajectory \\
    Average energy\linebreak expenditure & ${joule}$ &
    assuming unit mass, the robot energy expenditure is calculated as the integral of the squared robot velocity over the entire robot trajectory \\
    Average acceleration & $m/s^2$ & 
    average acceleration of the robot over its entire trajectory \\
    Average jerk & $m/s^3$ &
    average jerk (time derivative of acceleration) of the robot over its entire trajectory \\
    
    \addlinespace\rowcolor{Gray}
    \multicolumn{3}{c}{Pedestrian related} \\\addlinespace
    Closest-pedestrian\linebreak distance & $metres$ &
    at each robot trajectory segment, this is the distance to the closest-pedestrian \\
    Time-to-collision & $seconds$ &
    at each robot trajectory segment, this is the minimum time-to-collision for any pedestrian in the environment \\
    \bottomrule
    
    \end{tabular}
    \caption{The metrics suite implemented in \emph{SocNavBench}. See Sec.\ref{sec:metrics} for more details and considerations.}
    \label{tab:metrics}
\end{table}

We define several metrics that may be important considerations for a socially navigating robot and can help characterize the trade-off between robot appropriateness and efficiency.  
The metrics fall into four general evaluation categories:  path quality, motion quality, pedestrian disruption, and meta statistics.

\textbf{Path quality.}
Path quality metrics quantify the quality and efficiency of the path generated by the social navigation algorithm. These metrics are focused on the robot's path in a global sense.
    \begin{enumerate}
        \item Path length ${(meters)}$: the total distance traversed by the robot in the episode. Usually, lower is better and indicates that a robot takes more direct paths.
        \item Path length ratio: the ratio of straight line distance between the start and goal to the robot's  path length for any episode. Usually, higher is better and indicates that a robot takes more direct paths. Trade-offs between navigation optimality and pedestrian safety may affect this metric.
        \item Path irregularity ${(radians)}$: averaged over each point of the robot trajectory, this is the \added{absolute} angle between the robot heading and the vector pointing to goal \cite{guzzi2013human}. For a straight path from start to goal, path irregularity is zero. Usually, lower is better and indicates that a robot takes more direct paths, though as with path length ratio, this may vary based on other factors like pedestrian safety.
        \item Goal traversal ratio: calculated for \textit{Incomplete} episodes, this is the ratio of the robot's distance to goal at the episode \replaced{end to start}{start and end}. This is useful to quantify the partial success of algorithms when not completing the episode due to timeout or collision with the environment. Lower is better and indicates the robot \replaced{gets closer to the goal during \textit{Incomplete} episodes}{succeeds in its navigation more often}.
        \item Path traversal time $(seconds)$: total simulator time taken to traverse the robot's path. Lower is usually better but higher may be acceptable if the robot yields to pedestrians more frequently.
    \end{enumerate}
    
\textbf{Motion quality.}
The quality of the robot's motion is characterized by the smoothness and efficiency of the robot's movement. These metrics quantify the robot's energy expenditure, acceleration, and jerk. 
    \begin{enumerate}
        \item Average speed ${(meters/second)}$: average speed of the robot over its entire trajectory.
        \item Average energy expenditure ${(Joules)}$: robot energy is calculated as the integral of the squared robot velocity over the entire robot trajectory, assuming unit mass. Total robot energy expended is related to total path and velocity, but is slightly different from each. For example, two different algorithms may yield to pedestrians either by moving out of the way or by stopping before moving out of the way is necessary. The second algorithm may take the same overall time and have similar average velocity as the first but will expend lower energy in total. All other things equal, lower energy expenditure is better.
        \item Average acceleration ${(meters/second^2)}$: average acceleration of the robot over its entire trajectory.
        \item Average jerk ${(meters/second^3)}$: average jerk (time derivative of acceleration) of the robot over its entire trajectory. Lower indicates that the robot takes smoother paths, which are more predictable and legible for surrounding pedestrians while also consuming less energy. 
    \end{enumerate}

\textbf{Pedestrian-related.}
These metrics capture the robot's movements with respect to the surrounding pedestrians. They are mainly focused on the safety of the robot's navigation.
    \begin{enumerate}
        \item Closest-pedestrian distance $(meters)$: at each robot trajectory segment, this is the distance to the closest pedestrian. Pedestrian distances more than $10$ metres are saturated to $10m$.
        \item Time-to-collision $(seconds)$: at each robot trajectory segment, this is the minimum time-to-collision to any pedestrian in the environment. The time-to-collision \added{for any robot-pedestrian pair} is computed by \added{linearly} extrapolating the robot and pedestrian \added{trajectories} using their instantaneous velocity. \added{This is equivalent to the time it would take the robot-pedestrian pair to collide if, at the end of this trajectory segment, they continued moving at their current speed in their current heading. This is calculated for every robot-pedestrian pair in the environment and the minimum TTC is selected.} Times larger than $10$ seconds are saturated to $10s$.
    \end{enumerate}
    The two metrics above are similar but not the same and are better interpreted together. For example, a robot may follow a crowd walking in the same direction very closely, which would have a low \textit{Closest-pedestrian distance}. However, because their motions are in the same direction, the \textit{Time-to-collision} will also be high, indicating overall a lower pedestrian disruption. Between two algorithms with similar \textit{Time-to-collision}, users may prefer the higher \textit{Closest-pedestrian distance} in scenarios where giving pedestrians a wider berth is preferred. \added{These two metrics are range limited since instantaneous values of Times-to-collision greater than $10s$ and Closest-pedestrian distances greater than $10m$ are unlikely to impact the robot's operation in any way. Leaving these outliers in would affect their averages which may reflect an overall safer robot navigation than is actually the case.} 

\textbf{Meta statistics.}
Additionally, the tool generates a selection of overall success meta-statistics for the entire collection of episodes:
\begin{enumerate}
    \item Overall success rate: the fraction of episodes successfully completed (robot reached goal position and never collided with any pedestrian). Episodes may be completed but not successful if they have pedestrian collisions but no timeouts or environment collisions.
    \item Total pedestrian collisions: the total number of pedestrian collisions across all episodes considered. Lower is better.
    \item Failure cases: the distribution of failure cases over \textit{Timeout}, \textit{Pedestrian Collision}, and \textit{Environment Collision}. This is reported as a tuple (T/PC/EC). Lower numbers for each category are better.
    \item Average planning wait time $(seconds)$: this is the average time the simulator waited for commands from the planner. If the algorithms are tested on identical hardware, this provides a measure of the complexity of the algorithm and, hence, its feasibility for real-time, on-board robot operation. Lower is better.
\end{enumerate}
    
In the above descriptions, we describe the interpretation of metrics individually with all other metrics being equal. However, this is rarely the case during social navigation where algorithms are constantly negotiating a safety vs.\ efficiency trade-off. Hence, these metrics must be taken in context with each other, which is demonstrated in Sec.\ \ref{sec:exp_discussion}

\section{Experiments}
\subsection{Choice of candidate algorithms}
We evaluate three existing social navigation algorithms and a simple pedestrian-unaware baseline on \emph{SocNavBench}. 

We chose Social Forces \cite{helbing1995social}, ORCA \cite{vandenberg2008}, and SA-CADRL \cite{chen2017socially} as the three candidate social navigation algorithms because they represent the evolution of the field over three decades. The Social Forces model \cite{helbing1995social} is an early influential work on modeling pedestrian navigation that motivates many subsequent contributions in the robot navigation domain \cite{robicquet2016, zanlungo2011, moussaid2009}. ORCA \cite{vandenberg2008} is a popular baseline method used by many recent social navigation algorithms \cite{mavrogiannis2019effects, chen2017socially, chen2019} for performance comparisons. CADRL \cite{chen2017socially} is a recent, popular, and relatively state-of-the-art model using a deep reinforcement learning-based navigation method.


\textit{Social forces based pedestrian dynamics} (henceforth, Social Forces) is a formulation to explain pedestrian navigation behavior as being under the influence of "social forces" including an attractive force to their goal, a repulsive force from other pedestrians and obstacles, and other attractive forces to interesting objects or other pedestrians in the environment. We use an open-source implementation of this method from a popular pedestrian crowd simulator PedSim\cite{pedsim}. This method does not follow an angular velocity constraint.

\textit{ORCA: Optimal Reciprocal Collision Avoidance} \cite{vandenberg2008} (henceforth, ORCA) is an efficient navigation planner that provides theoretical non-collision guarantees assuming all agents in the environment use the same planning policy. The core idea behind reciprocal velocity obstacles used by ORCA is reciprocal reaction from both the pedestrian and the robot. Because in our simulator the pedestrians are not reacting to the robot, we changed the collaboration coefficients of the pedestrians to be 0 and that of the robot to be 1. However, ORCA, like most existing social navigation frameworks, does not account for system kinematics and dynamics in its formulation. Because its performance guarantees are void when a post-hoc kinematic constraint is applied, we run ORCA in an unbounded angular velocity mode without kinematic constraints on linear velocities.

\textit{SA-CADRL: socially aware collision avoidance with deep reinforcement learning} (henceforth, CADRL) \cite{chen2017socially} is a deep reinforcement learning based social navigation method which formulates social navigation as a cooperative multi-agent collision avoidance problem with well-crafted rules to inject social awareness. We are using the 4-agent version of the network. Because the number of agents to be fed into the network is fixed, we only take observations from the three nearest pedestrians to the robot. We used the pre-trained model provided by the authors, as described in the paper. Similar to ORCA, CADRL does not consider kinematic constraints.

Additionally, we implemented a meta-level planner for ORCA and CADRL because CADRL does not possess obstacle avoidance capabilities and the authors of CADRL and ORCA \cite{vandenberg2008crowdenvnav} endorsed the inclusion of a meta-level planner. The meta-level planner is responsible for identifying a near-future \replaced{waypoint}{checkpoint} for the robot to reach, directing the robot away from obstacles in the process. The planner we use is a sampling planner (from \cite{tolani2020visual}) that samples trajectories from a connectivity graph and evaluates them via heuristic cost functions and returns the minimum cost trajectory within a budget. Our meta planner has heuristic-based obstacle avoidance and goal-seeking capabilities. In operation, the meta planner plans a sub-goal within a 6-second horizon each time a new checkpoint is requested. This corresponds to around 4\textendash 7 meters. The robot is considered to have reached the checkpoint if it is within 1 meter of the checkpoint, and then a new checkpoint is requested.

\label{sec:baseline}
\textit{Pedestrian-unaware Baseline}: We use the meta-planner described above as a baseline navigation algorithm which does not take into account pedestrians around it and show its performance on the same set of navigation scenarios as the other 3 candidate algorithms. This planner can take into account kinematic constraints, so we apply a 3D (position and heading) unicycle model (Eq.\ \ref{eq:dubins}, Sec.\ \ref{sec:sim_bot_spec}). There is no kinematic constraint application on this planner when used as a meta-planner.

\subsection{Experimental protocol}
In our experiments, we used the default simulation settings, i.e. \textit{Synchronous} and \textit{Schematic} modes. In evaluation of algorithms that assume no kinematic constraints, we turned the kinematic constraints off, leaving only the maximum linear speed constraint for the robot.

In total, we tested on \added{the curated set of} $33$ episodes \added{described in Sec.\ \ref{sec:curated_eps}}. For the sake of easily comparing between different algorithms, we only consider those episodes in our analysis in which candidate algorithms complete the curated episodes. This is because if some algorithms terminate early due to pedestrian collisions, some metrics (such as path length or total energy expended) are not comparable. Hence, we did not terminate episodes for a pedestrian collision and instead counted all pedestrian collisions throughout the episode. $29/33$ episodes were completed by all $4$ candidate algorithms, and all metrics were computed on those $29$ completed episodes unless explicitly mentioned. 


\begin{table}[t]
    \centering
    \begin{tabular}{p{0.15\textwidth}
    p{0.07\textwidth}p{0.1\textwidth}p{0.12\textwidth}
    p{0.13\textwidth} 
    }
    \toprule
    \textbf{Candidate\linebreak Algorithm} 
    & \textbf{Overall success rate} & \textbf{Failure cases (T/PC/EC)} & \textbf{Total\linebreak pedestrian collisions$^+$} 
    & \textbf{Planning wall time per episode (s)$^+$}
    \\
    \midrule
    Social Forces~\cite{helbing1995social} &
    \textbf{32/33} & \textbf{(1/0/0)} & \textbf{1} & \textbf{{18.23 $\pm$ 7.41}} 
    \\\addlinespace[0.5em]
    
    ORCA~\cite{vandenberg2008} &
    24/33 & (1/8/0) & 15 & {48.84 $\pm$ 23.06}
    \\\addlinespace[0.5em]
    
    SA-CADRL~\cite{chen2017socially} &
    18/33 & (0/14/1) & 40 & {46.78 $\pm$ 21.38} 
    \\\addlinespace[0.5em]
    
    Baseline (S.\ \ref{sec:baseline}) &
    9/33 & (1/23/0) & 64 & {51.12 $\pm$ 16.21}
    \\
    \bottomrule
    \end{tabular}
    \caption{Average meta statistics for candidate algorithms tested on \emph{SocNavBench}. The metrics marked with $+$ are evaluated only on episodes completed by all algorithms.}
    \label{tab:metrics_meta}
\end{table}

\begin{table}[t]
    \centering
    \begin{tabular}{p{0.15\textwidth}
    p{0.11\textwidth}
    p{0.1\textwidth}
    p{0.08\textwidth}
    p{0.1\textwidth}
    p{0.11\textwidth}
    }
    \toprule
    \textbf{Candidate\linebreak Algorithm} 
    & \textbf{Path\linebreak length (m)} 
    & \textbf{Path\linebreak length ratio} 
    & \textbf{Goal\linebreak traversal ratio$^{-}$} 
    & \textbf{Path\linebreak irregularity (radians)}
    & \textbf{Path\linebreak traversal time (s)}
    \\
    \midrule
    Social Forces~\cite{helbing1995social}
    & {17.25 $\pm$ 4.05} 
    & {1.15 $\pm$ 0.23} 
    & 0.52
    & {1.66 $\pm$ 0.95} 
    & {15.93 $\pm$ 4.17} 
    \\\addlinespace[0.5em]
    
    ORCA~\cite{vandenberg2008} 
    & {17.66 $\pm$ 5.22} 
    & {1.17 $\pm$ 0.26}
    & 0.21
    & \textbf{{1.56 $\pm$ 1.01}}
    & {22.06 $\pm$ 7.78} 
    \\\addlinespace[0.5em]
    
    SA-CADRL~\cite{chen2017socially} 
    & \textbf{{15.70 $\pm$ 3.72} }
    & \textbf{{1.04 $\pm$ 0.05}}
    & 0.51
    & {1.68 $\pm$ 1.04} 
    & \textbf{{15.14 $\pm$ 4.21}}
    \\\addlinespace[0.5em]
    
    Baseline (S.\ \ref{sec:baseline}) 
    & {15.88 $\pm$ 3.57}
    & {1.05 $\pm$ 0.11} 
    & \textbf{0.09}
    & {1.65 $\pm$ 1.02} 
    & {16.08 $\pm$ 3.73} 
    \\
    \bottomrule
    \end{tabular}
    \caption{Average Path quality metric scores for candidate algorithms tested on \emph{SocNavBench}. The metrics marked with $-$ are evaluated only on incomplete episodes for each algorithm (T/EC).}
    \label{tab:metrics_path}
\end{table}

\begin{table}[ht]
    \centering
    \begin{tabular}{p{0.15\textwidth}
    p{0.12\textwidth}p{0.15\textwidth}
    p{0.15\textwidth}p{0.15\textwidth}
    }
    \toprule
    \textbf{Candidate\linebreak Algorithm} 
    & \textbf{Average\linebreak speed (m/s)\linebreak (max=1.2m/s)} & \textbf{Average energy\linebreak expenditure (J)}
    & \textbf{Average\linebreak acceleration (m/s$^2$)} & \textbf{Average\linebreak jerk (m/s$^3$)}
    \\
    \midrule
    Social Forces~\cite{helbing1995social} 
     & {1.09 $\pm$ 0.27} 
     & {398.33 $\pm$ 95.45} 
     & {0.39 $\pm$ 1.43} 
     & \textbf{{2.70 $\pm$ 28.23}  }   
    \\\addlinespace[0.5em]

    ORCA~\cite{vandenberg2008} 
    & {0.80 $\pm$ 0.27} 
    & \textbf{{315.03 $\pm$ 89.89}} 
    & {0.31 $\pm$ 1.26} 
    & {6.28 $\pm$ 28.85} 
    \\\addlinespace[0.5em]
    
    SA-CADRL~\cite{chen2017socially} 
    & {1.04 $\pm$ 0.36} 
    & {367.58 $\pm$ 87.24} 
    & {0.93 $\pm$ 2.91} 
    & {31.68 $\pm$ 87.71} 
    \\\addlinespace[0.5em]
    
    Baseline (S.\ \ref{sec:baseline})
    & {0.99 $\pm$ 0.42} 
    & {370.88 $\pm$ 84.55} 
    & {4.81 $\pm$ 9.07} 
    & {180.86 $\pm$ 303.88}     
    \\
    \bottomrule
    \end{tabular}
    \caption{Average Motion quality metric scores for candidate algorithms tested on \emph{SocNavBench}. 
    }
    \label{tab:metrics_motion}
\end{table}

\label{sec:results}
\begin{figure}[t]
    \centering
    \begin{subfigure}[b]{0.46\textwidth}
    \includegraphics[width=\textwidth]{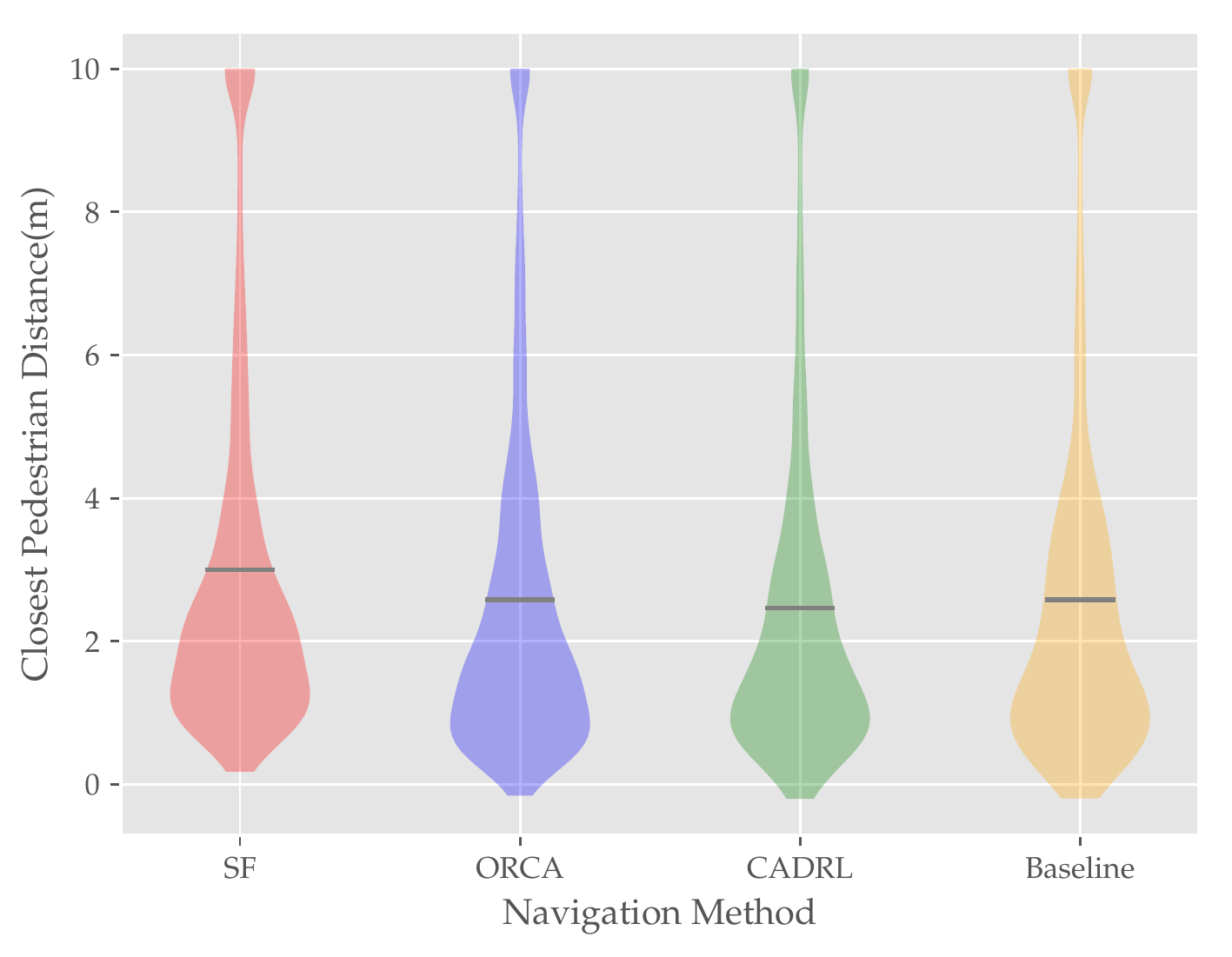}
    \caption{Closest-pedestrian distance}\label{fig:cpd}
    \end{subfigure}
\begin{subfigure}[b]{0.46\textwidth}        
    \includegraphics[width=\textwidth]{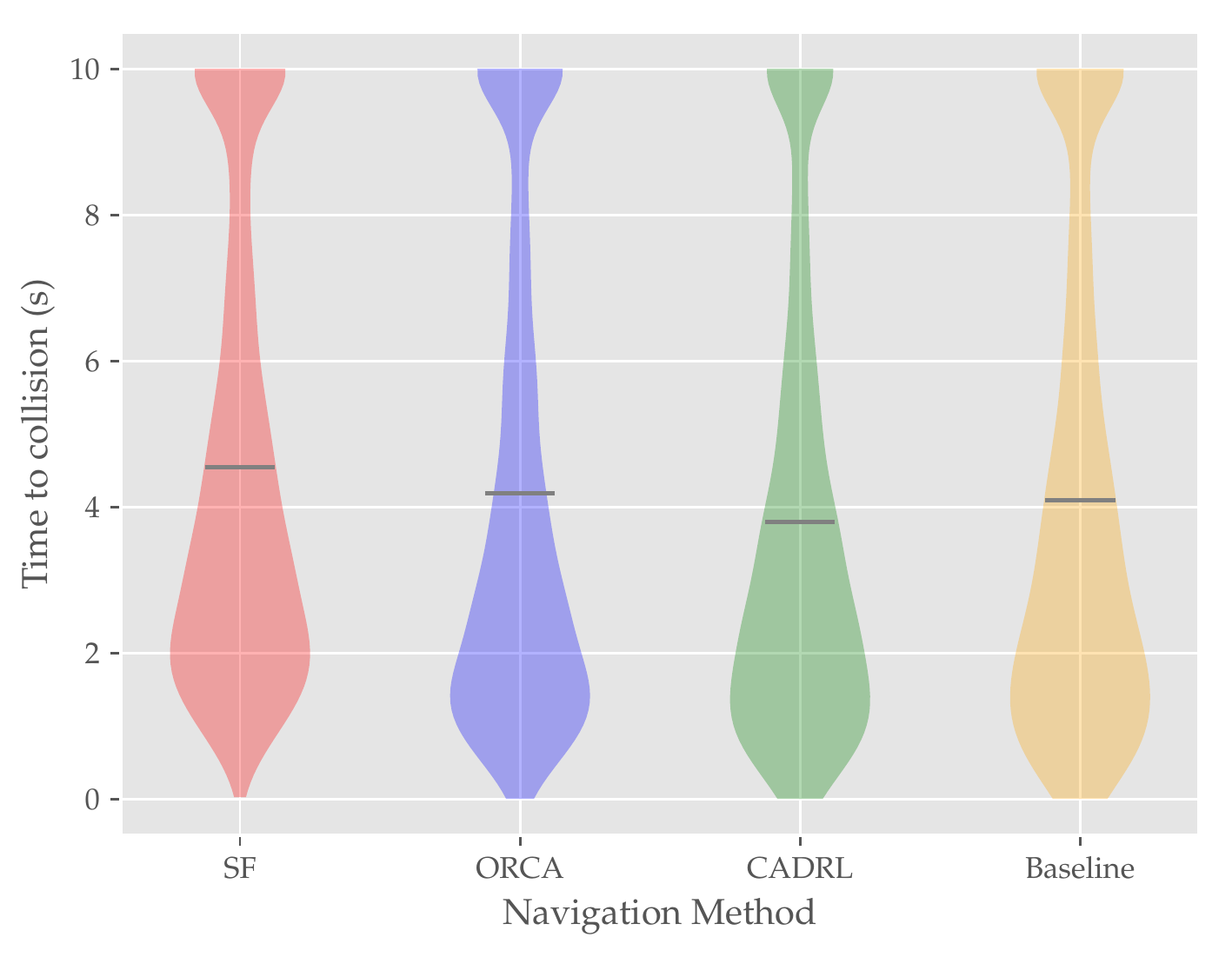}
    \caption{Time to collision}\label{fig:ttc}
    \end{subfigure}
    \caption{Pedestrian-related statistics for the candidate algorithms. Gray lines indicate the distribution mean. CPD is saturated beyond $10m$ and TTC beyond $10s$ which results in bimodality towards higher values CPD and TTC instead of a (very) long tail due to outliers.}\label{fig:pedstats}
\end{figure}


\begin{figure}[t]
    \centering
    \begin{minipage}[t]{.48\linewidth}
    \includegraphics[width=\textwidth]{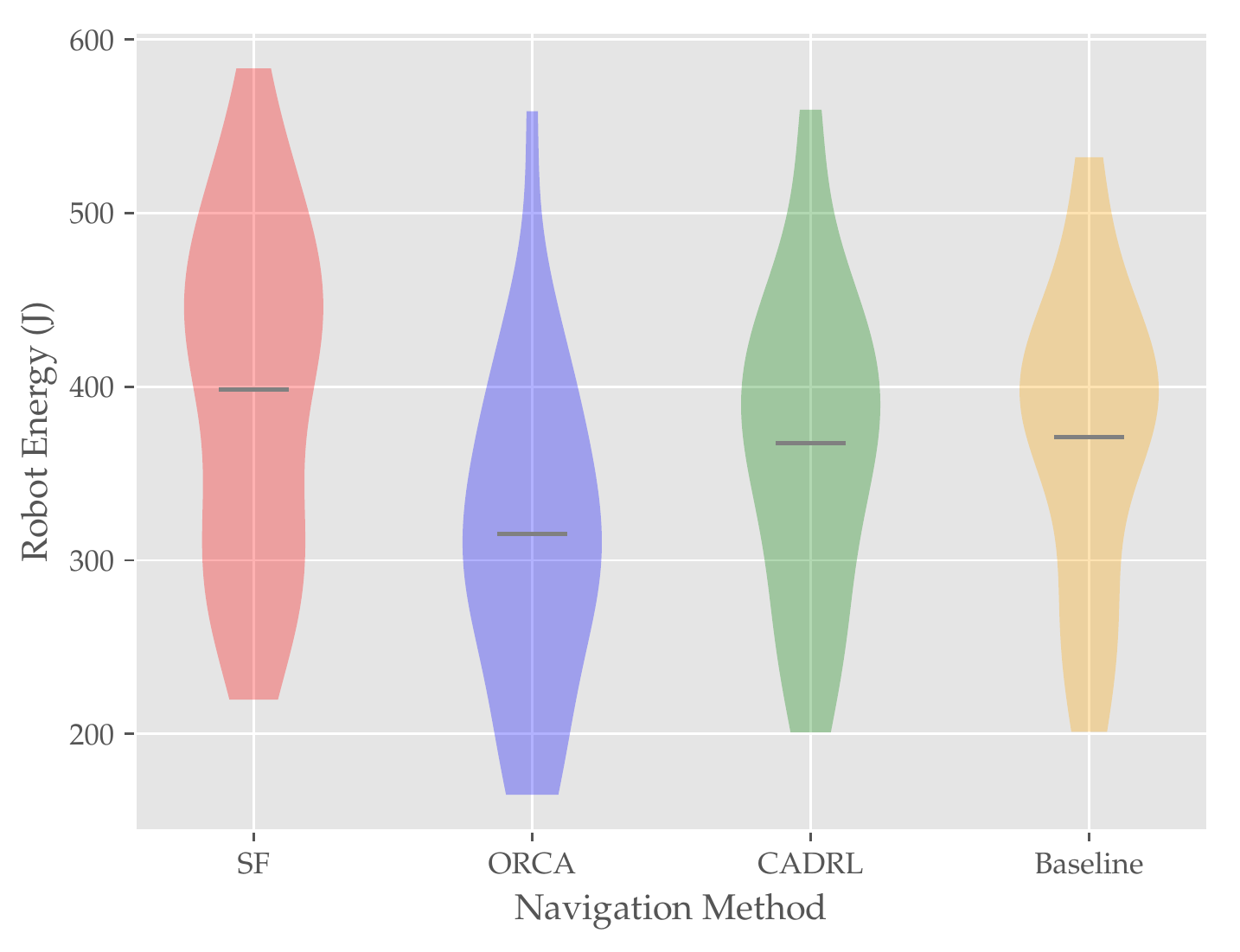}
    \caption{Average robot motion energy expenditure per episode.}\label{fig:energy}
    \end{minipage}\hfill
    \begin{minipage}[t]{.48\linewidth}
    \includegraphics[width=\textwidth]{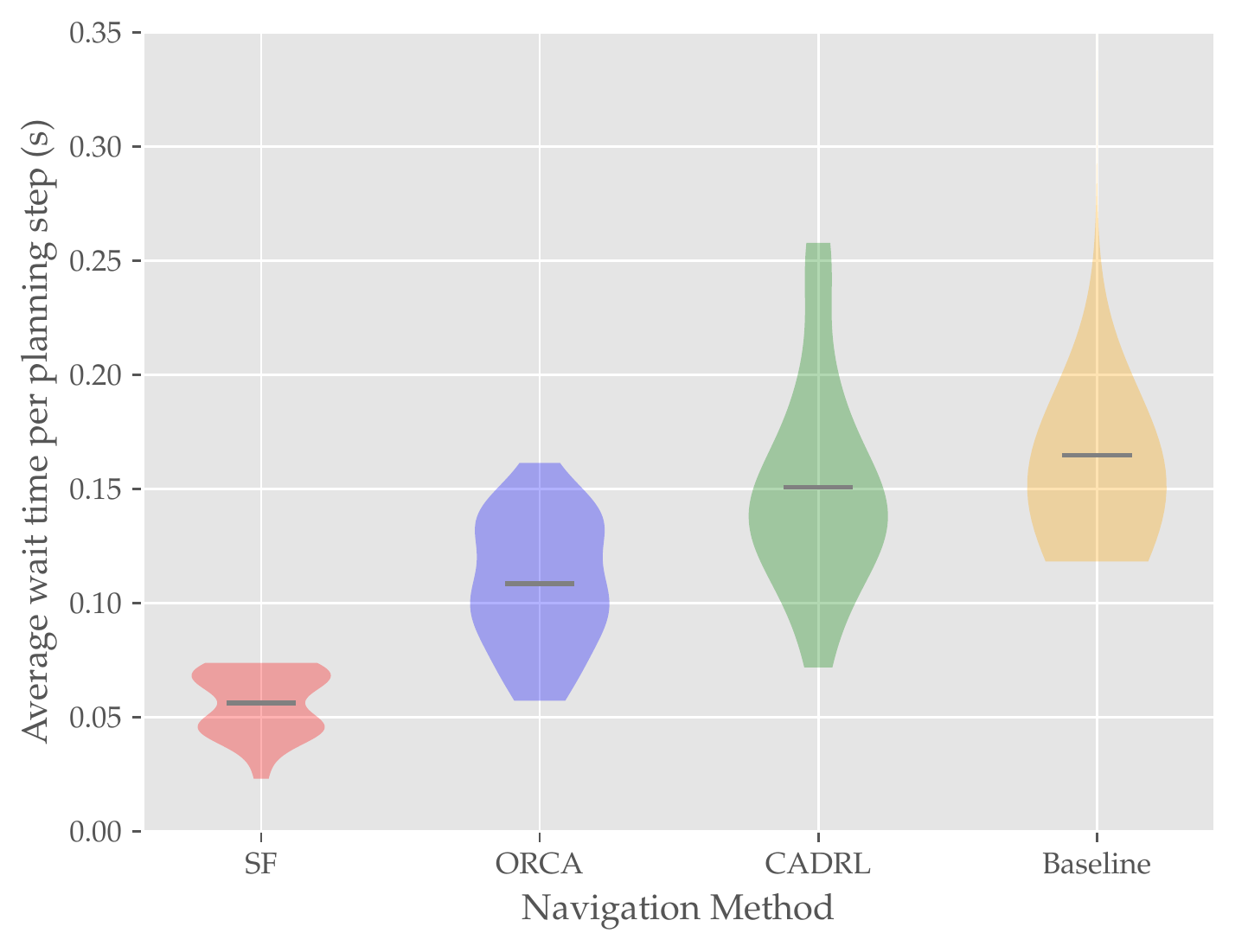}
    \caption{Average wall-clock wait time per planning step.}\label{fig:wwt}
    \end{minipage}
\end{figure}

\begin{figure}[t]
    \centering
    \begin{minipage}[t]{.32\linewidth}
    \includegraphics[width=\textwidth]{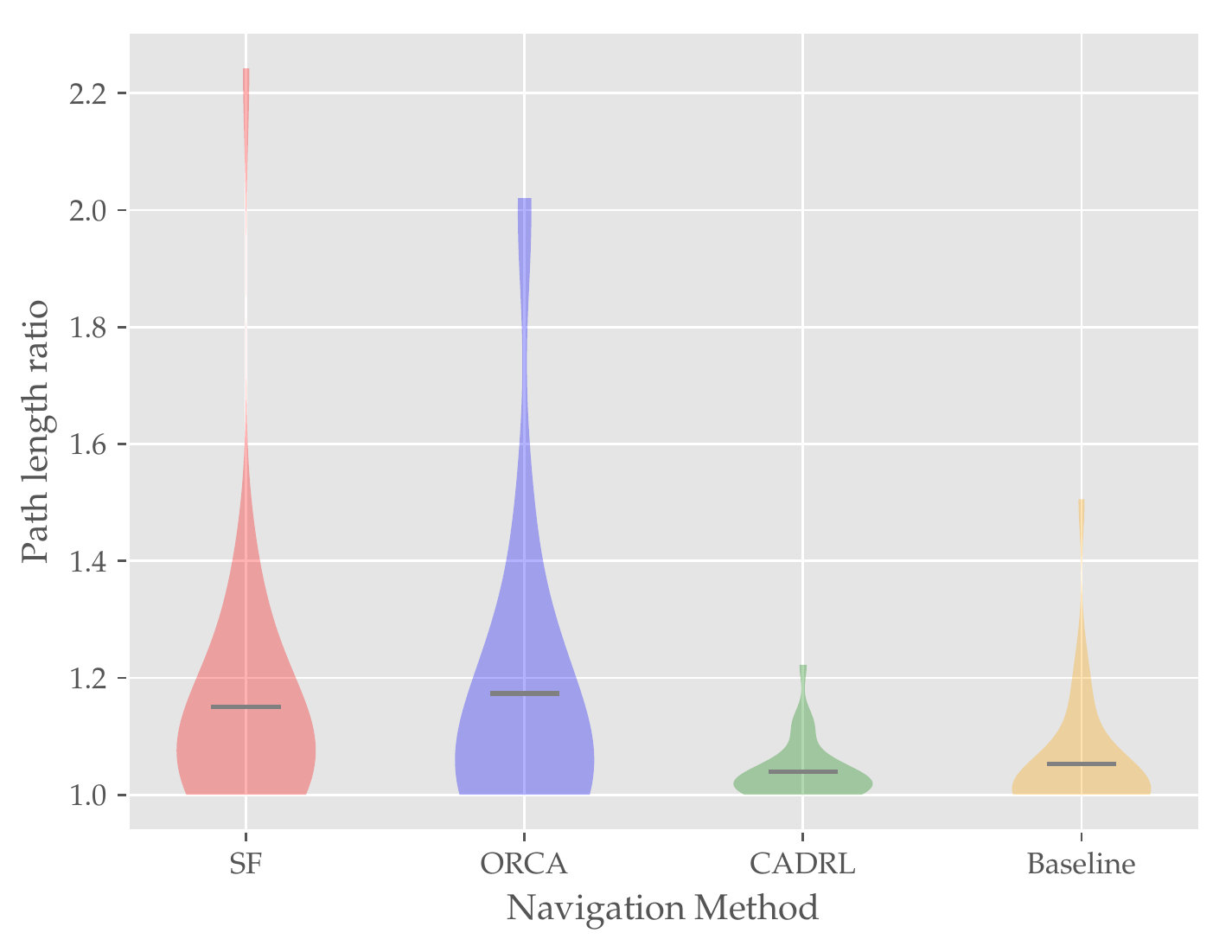}
    \caption{Average path length ratio per episode.}\label{fig:path_length_ratio}
    \end{minipage}\hfill
    \begin{minipage}[t]{.32\linewidth}
    \includegraphics[width=\textwidth]{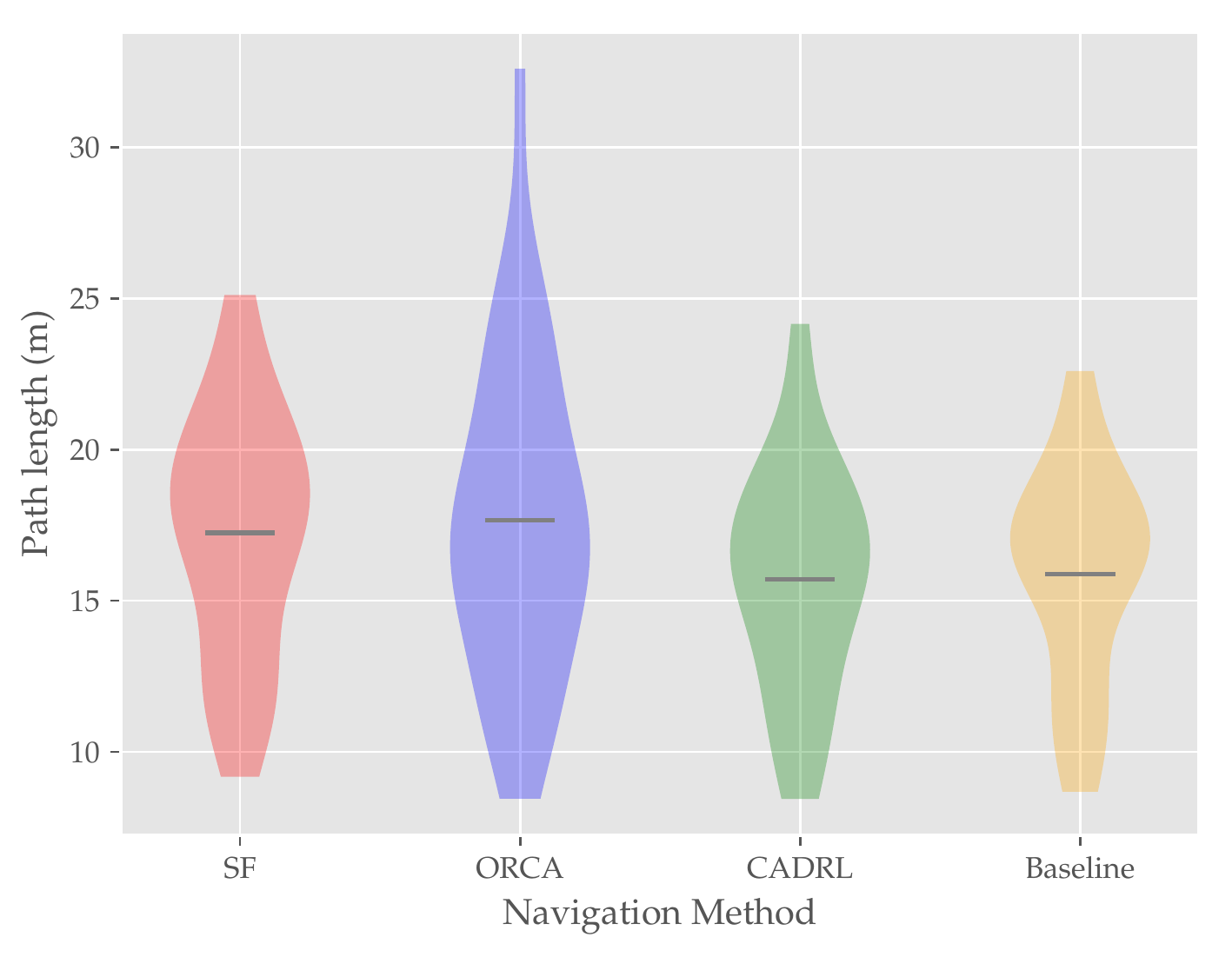}
    \caption{Average path length per episode.}\label{fig:path_length}
    \end{minipage}
    \begin{minipage}[t]{.32\linewidth}
    \includegraphics[width=\textwidth]{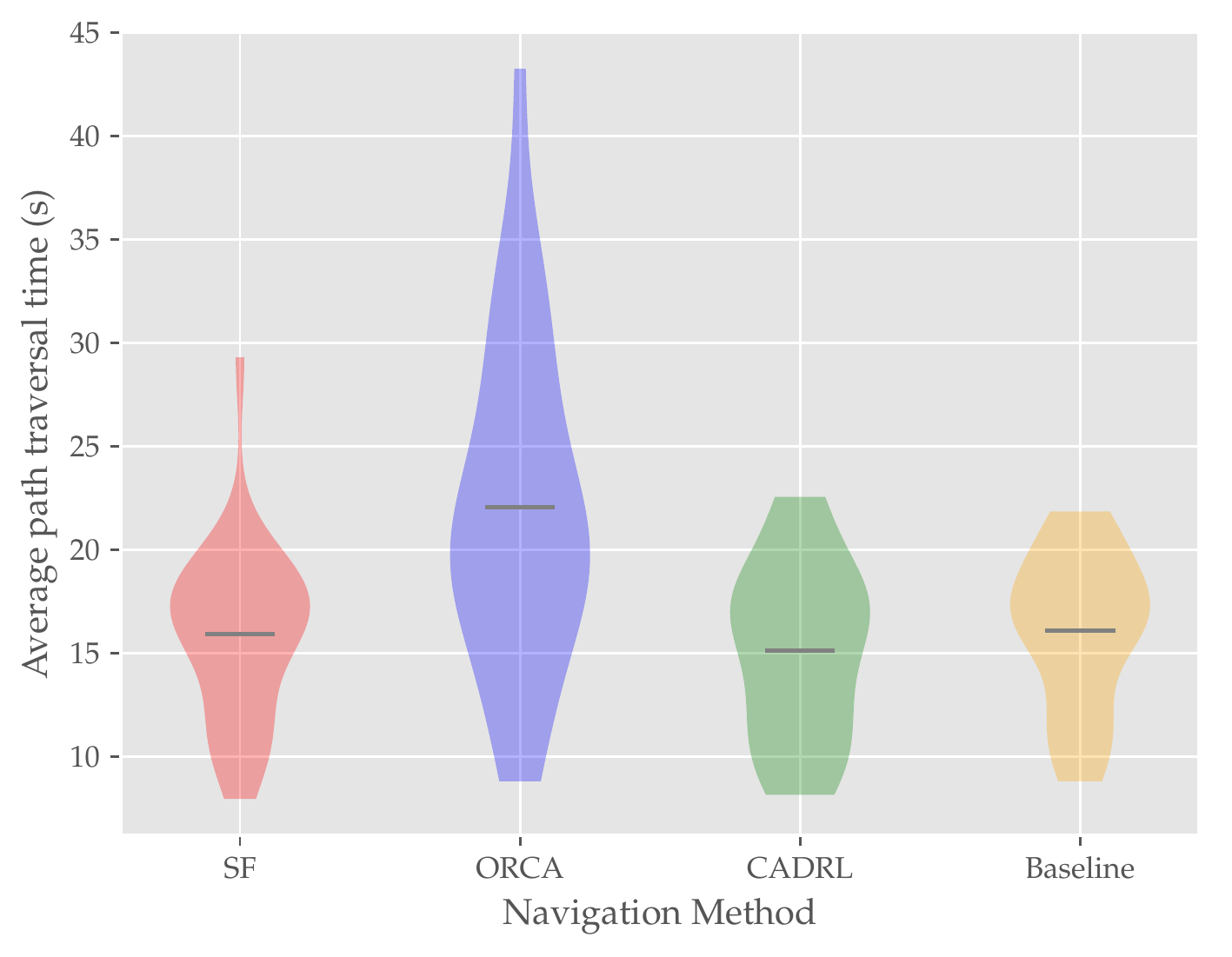}
    \caption{Average path traversal in simulator time per episode.}\label{fig:tstt}
    \end{minipage}
\end{figure}

\section{Results \& Discussion}
We ran the aforementioned social navigation methods on the curated set of episodes described in Sec.\ \ref{sec:curated_eps} and discuss the results below. Our focus is on how each metric reflects the overall navigation style of the algorithm with a particular focus on the preferences inherent in the algorithms with respect to the navigation efficiency versus pedestrian disruption trade-off. \added{Figures \ref{fig:pedstats} - \ref{fig:tstt} are violin plots which represent the shape of the distribution of each metric, with the mean of each distribution marked with a horizontal gray tick on the violin. This helps us see outliers and modalities in addition to the information provided by error bars.}

\subsection{Experimental results}
\label{sec:exp_discussion}
\subsubsection{\textbf{Meta statistics} (Table\ \ref{tab:metrics_meta})}\par

From Table\ \ref{tab:metrics_meta}, the Social Forces model is the most successful social navigation method with the highest success rate $32/33$ and no pedestrian collisions. Hence, all completed episodes were also successful. It also has the fastest planning time, largely because it does not require a meta-planner.

ORCA is the second most successful model with $24$ successful cases out of the total $33$ and $15$ pedestrian collisions across $8$ episodes. It completed $32/33$ episodes. The planning time per step for ORCA is close to the baseline (Table~\ref{tab:metrics_meta} \& Fig.~\ref{fig:wwt}), indicating that the computation bottleneck is the meta-planner.

SA-CADRL, despite being relatively state-of-the-art, frequently runs into pedestrians with $40$ collisions across $14$ episodes. Its overall completion rate is, however, the same as ORCA. The testing scenarios in the curated dataset are challenging, with sometimes more than 3 pedestrians moving relatively close to the robot. It is very likely that the robot frequently runs into situations unseen during training and causes collisions. Similar to ORCA, the planning time is close to the baseline, indicating the same computation bottleneck.

Unsurprisingly, the pedestrian-unaware baseline performs the worst with respect to pedestrian safety. While it did complete all but one episode by reaching the goal, it collided with a total of $60$ pedestrians across $23$ episodes.

Additionally, all candidate algorithms had one \textit{Incomplete} episode each, which were all distinct episodes. This illustrates the diversity and difficulty of our curated set of episodes, which test different aspects of social navigation.

\added{Meta-statistics reflect an overall reliability and computation time for each model. From Table\ \ref{tab:metrics_meta}, the Social Forces model is computationally fastest and most reliable model because it passes the most test scenarios with the fewest pedestrian collisions.}

\subsubsection{\textbf{Path quality metrics} (Table\ \ref{tab:metrics_path})}\par

The Social Forces model takes very circuitous paths to goal, as reflected in long path lengths (Fig.\ \ref{fig:path_length}) and high path length ratios (Fig.\ \ref{fig:path_length_ratio}). However, judging by its high success rate, we can infer that it is being conservative about avoiding pedestrians. This conservative behavior is further demonstrated by high path irregularity, as it tends to take large curved paths. Despite having long path lengths, the Social Forces model has low path traversal time, meaning it is consistently guiding the robot to move at full speed.

ORCA has the longest path lengths and the second highest success rate, which would seem to imply that ORCA also makes a significant effort to avoid pedestrians by taking roundabout paths. However, given that ORCA also has the lowest path irregularity, we can infer that it is actually trying to take relatively direct paths to goal and using local deviations while trying to to avoid pedestrians.
ORCA's longest path traversal times also imply that ORCA is taking a greedy approach of heading towards the goal and using deviations while avoiding pedestrians along the way. A qualitative analysis of the trajectories, paired with the metrics data, indicates that ORCA tends to react late to pedestrians during avoidance maneuvers and then buy time for an avoidance maneuver by starting to move in the opposite direction of the pedestrian (Fig. \ \ref{fig:qual_algos}).
This greedy approach often runs into problems when it is too late to react.

CADRL has the least path length and path traversal time, indicating aggressive goal-seeking behavior. 

The baseline method has low path lengths overall, just behind CADRL. This is explained by the fact that the baseline method is subject to kinematic constraints via a unicycle model, which limits its liner and angular acceleration capabilities. It has a larger turning radius than the other candidate algorithms, which is indicated by its high path irregularity.


\added{The path quality metrics mostly reflect a model's efficiency while path irregularity and path length ratio act as sanity checks for presence of large detours or frequent zigzagging patterns. We can see that SA-CADRL is the fastest and the most efficient model overall. However, its aggressive evasive maneuvers would mean a high path irregularity. ORCA produces paths that are direct with less aggressive evasive maneuvers at the cost of efficiency. The navigation driven by Social Forces tended to take large detours.}

\subsubsection{\textbf{Motion quality metrics} (Table\ \ref{tab:metrics_motion})}\par
The inferences we made about the algorithms from the previous sections are further confirmed by the results in Table\ \ref{tab:metrics_motion}.

Social Forces has the fastest average speed. This confirms that despite having long paths, it is operating at almost maximum speeds, consistently leading to low path traversal times (Table~\ref{tab:metrics_path}). The smoothness of motion is further illustrated by low robot acceleration and jerk. Due to its behavior being conservative and its relatively longer paths traversed with high speeds, it also happens to consume the highest amount of energy.

ORCA takes relatively direct approaches to the goal, as shown by consuming the least amount of energy here. Low acceleration and jerk indicate that its motion is relatively smooth as well, albeit at not as smooth as the Social Forces model. This implies that ORCA's reactions to pedestrians are not very sudden or unpredictable.

CADRL also is very fast, which is again explained by its aggressiveness in goal seeking. This aggressiveness is further shown by high acceleration and jerk, which indicate sudden movements to dodge pedestrians. Despite being aggressive in proceeding towards the goal, it expends plenty of energy, most of which is spent on accelerating and decelerating the robot.

\added{The motion quality metrics reflect how predictable a model's produced robot actions are. We can see that SA-CADRL has large acceleration and jerk. Therefore, coupled with its highly direct paths, it is recommended over the other models in situations where the robot has fewer energy constraints such as small operating range. The Social Forces model produces the smoothest trajectories, while ORCA closely follows.}

\subsubsection{\textbf{Pedestrian disruption metrics} (Figure  \ref{fig:pedstats})}\par

We can see again from Fig.~\ref{fig:pedstats} that the Social Forces model is the most conservative and safest model because it has the furthest closest-pedestrian distances with no collisions and longest times-to-collision. ORCA tends to avoid pedestrians locally and will run into clusters of pedestrians hoping to find gaps, so it has a smaller closest distance and short times-to-collision. CADRL is very aggressive and only dodges pedestrians at the last second. This is shown by the shortest closest distance and short times-to-collision. 
This observation of CADRL's aggressiveness was also made in prior work \cite{chen2019}.  
For CADRL and ORCA, the closest-pedestrian distance has some negative values because there are episodes with collisions between the simulated robot and pedestrians. \added{Pedestrian disruption metrics indicate the navigational compliance of a robot by measuring how much it avoids or yields to pedestrians. We can infer that Social Forces is the most compliant model, because of its high minimum distance to pedestrians and large times to collision. Conversely, SA-CADRL is the most aggressive model, with ORCA in between.}

\subsection{Summary of candidate algorithm performance}
\label{sec:qual_discussion}

\begin{figure}[t]
    \centering
    \includegraphics[width=\textwidth]{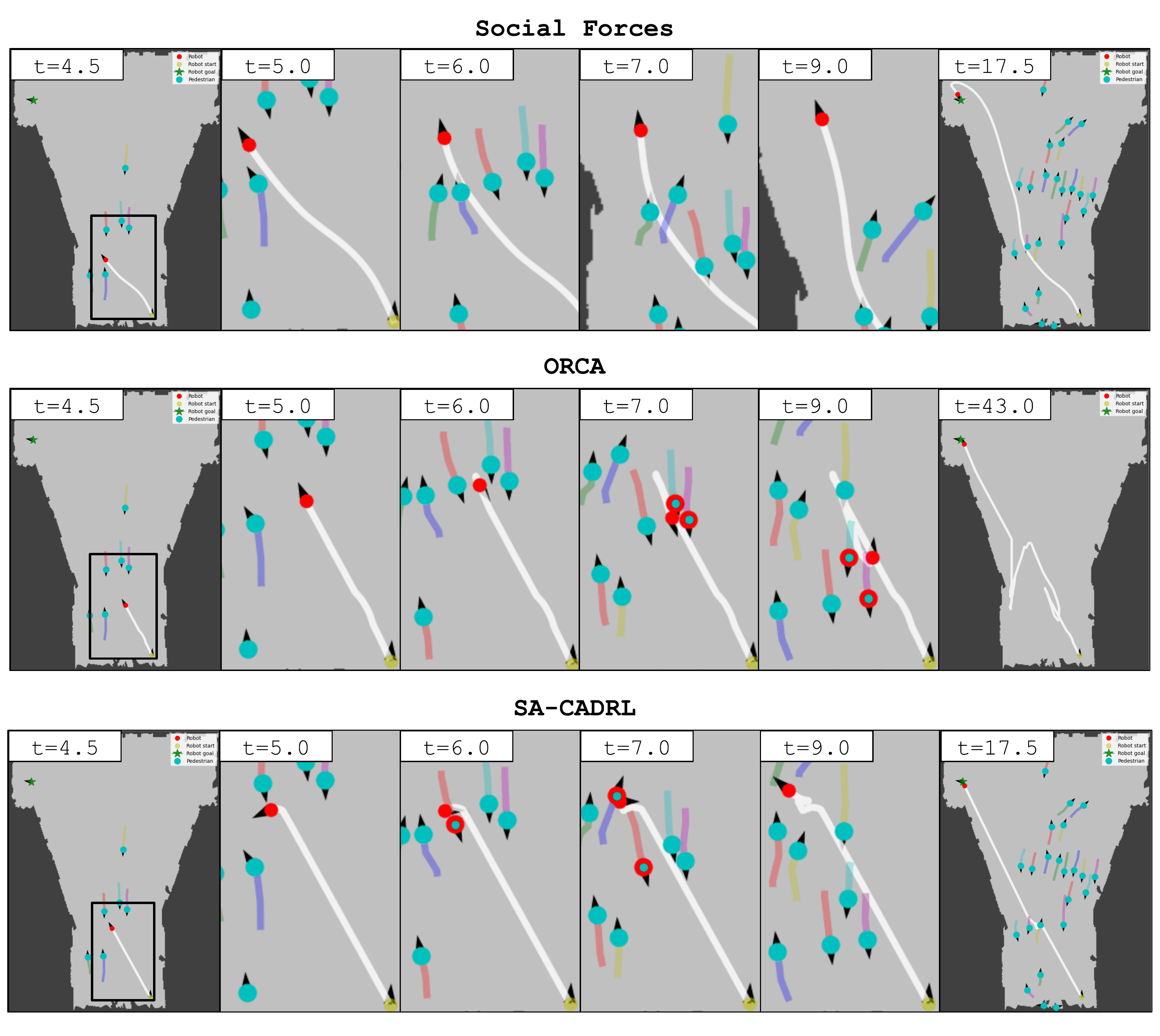}
    \cprotect
    \caption{Qualitative comparisons for candidate algorithms' pedestrian crossing behavior tested on \emph{SocNavBench}. The leftmost image shows the simulated robots about to cross a dense pedestrian crowd to get to their goal. The rightmost image shows the simulated robots just about to reach the goal. Pedestrians with whom collisions occurred are indicated with a red border. A discussion of the algorithms' behavior can be found in Section \ \ref{sec:qual_discussion}.}
    \label{fig:qual_algos}
\end{figure}

Overall, the Social Forces model produces robot motion that errs on the side of caution. In our testing, it produced long paths that take care to avoid pedestrians but also traversed them quickly. It was also the computationally quickest planner amongst all tested algorithms, although this is partly because it did not require a meta-level planner for obstacle avoidance. It also suffers from an issue where a large cluster of pedestrians away from the goal can cause it to overshoot the goal due a large repulsive force. This can be seen in the last frame for Social Forces in Fig. \ \ref{fig:qual_algos}, where it overshoots and loops around the goal. It also had one failure due to \textit{Timeout} in which it was stuck in a local minimum because of the relative configuration of the goal position and the environmental obstacles (it was unable to navigate around a corner to a goal). 


ORCA tried to strike a middle ground between the Social Forces model and CADRL in terms of goal-seeking aggressiveness. Its pedestrian avoidance was more locally focused than Social Forces and similar to SA-CADRL in that it tried to deal with individual pedestrians closest to its path rather than planning around pedestrians or groups of pedestrians as a whole. Most often, the local avoidance behaviors amounted to the robot trying to escape pedestrians by straying from its preferred trajectory. This behavior can be observed in Fig. \ \ref{fig:qual_algos}. At $t=5$, the robot is still proceeding towards goal, but turns around and starts to move away from the approaching pedestrians once they get too close. However, it reacts too late, and ends up with colliding with multiple pedestrians. This behavior is repeated with subsequent groups of approaching pedestrians as can be seen from the path history and long total time taken as indicated in the last frame.
Once the pedestrians had passed and were no longer on the robot's subsequent path segments, it started to plan towards the goal.

SA-CADRL was the most aggressive of the 3 social navigation models in terms of goal-seeking behavior. It generated direct paths to goal and deals with pedestrians via local avoidance behaviors. Most often, it would stop and turn away from pedestrians but would not move to accommodate them if they kept moving on their trajectory (Fig. \ \ref{fig:qual_algos}). \added{Therefore in situations where the robot takes precedence over surrounding pedestrians while still trying to avoid them, such as when ferrying emergency supplies through a crowded environment, SA-CADRL would be an appropriate choice due to its direct and efficient paths which still have some pedestrian avoidance capabilities. In contrast, if the robot takes less precedence than pedestrians who share space with it (such as a laundry collecting robot in a hospital), Social Forces may be an appropriate model since it yields to other agents with higher priority (such as healthcare workers or patients with limited mobility)}

\added{Our findings are partly corroborated by previous work. The experiments conducted by the authors of SA-CADRL \cite{chen2017socially} showed, in agreement with our findings, that SA-CADRL is faster than ORCA in terms of path traversal time. However, they showed that SA-CADRL has shorter minimum separation distance to pedestrians (analogous to our \textit{Closest pedestrian distance}) while we observe similar measurements in this metric (see Fig.~\ref{fig:pedstats}(a)). One difference between our experiment and theirs is that they used a fixed two or four-agent setting where every agent follows the same policy whereas our agent is considered invisible to the real world pedestrians. Apart from traversal time and minimum separation distance, no additional metrics are mentioned in this work. In terms of the Social Forces policy, some works highlighted its fast computation time \cite{Sud2008} and high success rates \cite{ferrer2013} similar to what we observed, but to the best of our knowledge, no prior work evaluated the Social Forces model in the same experimental setting as ORCA or CADRL. The lack of a descriptive suite of metrics and comparisons across different policies in a formal setting further highlights the importance of our benchmark.}

\subsection{Considerations around pedestrian trajectory replay}
\label{sec:replay_assumptions}
\added{
SocNavBench uses a replay style simulation, meaning that simulated pedestrians execute pre-recorded trajectories and are not responsive to the robot's actions. This replay style method introduces some limitations to the benchmark that we believe are mitigated in many scenarios. In this section, we describe our assumptions about scenarios where SocNavBench works well, and detail the limitations of our benchmark. Additionally we  contrast the tradeoff between our approach and a reactive pedestrian simulation.

The key assumption that underlies SocNavBench metrics is that interrupting pedestrians by inducing deviations from their preferred trajectory is a sign of poor social navigation.  For example, the  distribution of the  \textit{Time-to-collision} metric (Fig. \ref{fig:ttc}) quantifies how much the navigation algorithm puts the robot in ``close to collision'' scenarios. Algorithms that produce consistently small times-to-collision have a higher likelihood of inducing responsive humans to deviate from their originally preferred trajectory, and thus rank less well.


We believe the aforementioned assumption holds for pedestrians in larger, densely crowded areas where typically flows of pedestrians emerge, such as a large atrium or wide sidewalk. 
In such a situation, it would be less socially appropriate for a single agent (robot or human) to disrupt these emergent pedestrian flows by cutting off pedestrians or trying to ``shoot the gap'' and failing. 
Since our benchmark and curated episodes involves large outdoor maps with an average of $44$ pedestrians per episode, this assumption is appropriate for the scenarios described in \textit{SocNavBench}. 

In contrast, consider a scenario involving a single pedestrian in a supermarket aisle or hospital hallway. This individual pedestrian may walk down the middle of the aisle in the absence of a second agent, but it is still socially acceptable for them them to move closer to one side to allow an opposing agent to pass. In this case, the robot inducing a deviation in the pedestrian's motion is not a sign of poor social navigation; on the contrary, it reflects effective negotiations of space. In these scenarios, SocNavBench metrics do not capture the social aspects of navigation well. 

Our non-reactive agent assumption can still be useful for some real-world situations in which reactive modeling may seem imperative. Consider smaller environments, such as the aforementioned hospital scenario---if the other agents in the scene take much higher precedence over the robot, our approach of minimizing deviations of others at the expense of the robot is still useful. Such a precedence order would be common in a hospital where a robot such as the TUG~\cite{TUGbot} ferries laundry and other non-critical supplies between locations. In this situation, it would be undesirable for the TUG bot to interrupt the paths of critical operations such as care providers~\cite{mutlu2008robots}, patients being moved, or even individuals for whom path changes are expensive, such as wheelchair users and assisted walkers.


We chose not to include pedestrian reactivity in this version of SocNavBench mainly because reactivity adds a different set of limitations, mainly in the form of introducing biases. To incorporate reactive pedestrians, we would need to use a pedestrian movement model that can react to a simulated robot. Since no perfect pedestrian model exists, any model will include assumptions about the nature of pedestrian walking policies, introducing biases on various axes including  preferred speeds, accelerations, acceptable closeness to others, etc. Mitigating these biases requires considerable human data and analysis, which would require a full, independent research project. Additionally, such a benchmark would be prescriptive of a particular type of pedestrian navigation as extensible to all situations. Instead \textit{SocNavBench} is descriptive and the metrics herein can be interpreted as appropriate for the particular navigation context of candidate algorithms.
}

\added{Given the above considerations, we understand we cannot fully simulate the behavior of real \textit{reactive} humans, so real-world experiments will still be necessary to provide a complete picture of social navigation performance. Hence, \emph{SocNavBench} can serve as a complementary tool to assist in the development of social navigation algorithms by providing an interpretable performance benchmark that is cheaper to run and more consistent across algorithms than experiments in the real world, while being representative of real world pedestrian behavior --- which is a well accepted paradigm in human-robot interaction \cite{steinfeld2009ozofwizard}. This will also allow users to identify which parameters and algorithms to select prior to a real-world evaluation.}
%

\added{Although navigation scenarios in which it is important to model pedestrian reactivity are not well represented in \textit{SocNavBench} as presently constructed, the benchmark is built to be extensible to these scenarios. SocNavBench has an object-oriented structure and we provide a base \texttt{Human} class which is extended by a \texttt{PrerecordedHuman} class to achieve our pedestrian replay. The \texttt{update} methods of this class can be overridden to support reactive pedestrians, if researchers concerned with a particular type of pedestrian behavior have a policy model for it.}

\subsection{Future work} 
\label{sec:future}

 In the future, we plan to add an option to enable partial reactivity in the pre-recorded pedestrians so that they can react to imminent collisions by \replaced{switching to a \textit{situationally appropriate} pedestrian motion model with the same goal as the original pre-recorded pedestrian}{pausing their motion}. \added{This will allow us to add metrics to our evaluation suite that measure the number of times a particular social navigation algorithm causes pedestrian route changes as well as the amount of deviation caused. For added flexibility and ease of use in robot simulation, we also plan to provide a direct ROS interface via ROS messages (current sockets allow python and C++ ROS nodes to interact with \textit{SocNavBench} via native types) as well as support for custom robots through a URDF specification.}




\section{Conclusion}

We present \emph{SocNavBench} an integrated testing and evaluation framework for social navigation algorithms. \emph{SocNavBench} comprises a photorealistic simulator and a curated set of social navigation scenarios based on real-world pedestrian trajectory data on which social navigation algorithms can be tested. We also provide an implementation of a suite of metrics that are useful for evaluating the performance of these algorithms along various criteria including robot navigation efficiency and pedestrian disruption. 
Finally, we demonstrate the use of \emph{SocNavBench} by evaluating a set of three existing social navigation methods. \emph{SocNavBench} is publicly available at \href{https://github.com/CMU-TBD/SocNavBench}{this url}.

\section{Acknowledgements}

\added{We thank the anonymous reviewers of this work who helped improve the paper in several areas including clarity surrounding technical details and the arguments around the limits of pedestrian trajectory replay.} This work was funded under grants from the National Science Foundation (NSF IIS-1734361) and the National Institute on Disability, Independent Living, and Rehabilitation Research (NIDILRR 90DPGE0003).

\bibliographystyle{ACM-Reference-Format}
\bibliography{soc_nav}

\appendix

\end{document}